\documentclass[9pt,a4paper,english]{IEEEtran}
\usepackage[T1]{fontenc}
\usepackage{verbatim}
\usepackage{float}
\usepackage{amsbsy}
\usepackage{amstext}
\usepackage{amsthm}
\usepackage{graphicx}

\makeatletter

\floatstyle{ruled}
\newfloat{algorithm}{tbp}{loa}
\providecommand{\algorithmname}{Algorithm}
\floatname{algorithm}{\protect\algorithmname}

\theoremstyle{plain}
\newtheorem{thm}{\protect\theoremname}
\theoremstyle{plain}
\newtheorem{prop}[thm]{\protect\propositionname}

\usepackage{amsmath}
\usepackage{amssymb}
\usepackage[numbers,sort]{natbib}

\usepackage{graphicx}
\usepackage{subfigure}
\usepackage{pdfsync}
\usepackage{url}
\usepackage{pdfsync}
\usepackage{todonotes}
\makeatother

\usepackage{babel}
\providecommand{\propositionname}{Proposition}
\providecommand{\theoremname}{Theorem}

\begin{document}

\title{Successive Convex Approximation Algorithms for Sparse Signal Estimation
with Nonconvex Regularizations}

\author{Yang Yang, Marius Pesavento, Symeon Chatzinotas and Björn Ottersten\thanks{Y. Yang, S. Chatzinotas and B. Ottersten are with Interdisciplinary
Centre for Security, Reliability and Trust, University of Luxembourg,
L-1855 Luxembourg (email: yang.yang@uni.lu, symeon.chatzinotas@uni.lu,
bjorn.ottersten@uni.lu). Their work is supported by the ERC project
AGNOSTIC.}\thanks{M. Pesavento is with Communication Systems Group, Technische Universit\"{a}t
Darmstadt, 64283 Darmstadt, Germany (email: pesavento@nt.tu-darmstadt.de).
His work is supported by the EXPRESS Project within the DFG Priority
Program CoSIP (DFG-SPP 1798).}}
\maketitle
\begin{abstract}
In this paper, we propose a successive convex approximation framework
for sparse optimization where the nonsmooth regularization function
in the objective function is nonconvex and it can be written as the
difference of two convex functions. The proposed framework is based
on a nontrivial combination of the majorization-minimization framework
and the successive convex approximation framework proposed in literature
for a convex regularization function. The proposed framework has several
attractive features, namely, i) flexibility, as different choices
of the approximate function lead to different type of algorithms;
ii) fast convergence, as the problem structure can be better exploited
by a proper choice of the approximate function and the stepsize is
calculated by the line search; iii) low complexity, as the approximate
function is convex and the line search scheme is carried out over
a differentiable function; iv) guaranteed convergence to a stationary
point. We demonstrate these features by two example applications in
subspace learning, namely, the network anomaly detection problem and
the sparse subspace clustering problem. Customizing the proposed framework
by adopting the best-response type approximation, we obtain soft-thresholding
with exact line search algorithms for which all elements of the unknown
parameter are updated in parallel according to closed-form expressions.
The attractive features of the proposed algorithms are illustrated
numerically.
\end{abstract}

\begin{IEEEkeywords}
Big Data, Line Search, Majorization Minimization, Nonconvex Regularization,
Successive Convex Approximation
\end{IEEEkeywords}

\section{Introduction}

In this paper, we consider the following optimization problem
\begin{align}
\underset{\mathbf{x}}{\textrm{minimize}}\quad & h(\mathbf{x})\triangleq f(\mathbf{x})+g(\mathbf{x}),\label{eq:general-formulation}
\end{align}
where $f$ is a smooth function and $g$ is a nonsmooth function.
Such a formulation plays a fundamental role in parameter estimation,
and typically $f$ models the estimate error while $g$ is a regularization
(penalty) function promoting in the solution a certain structure known
a priori such as sparsity \cite{Theodoridis_2015}. Among others,
the linear regression problem is arguably one of the most extensively
studied problems and it is a special case of (\ref{eq:problem-formulation})
by setting $f(\mathbf{x})=\frac{1}{2}\left\Vert \mathbf{Ax-y}\right\Vert _{2}^{2}$
and $g(\mathbf{x})=\lambda\left\Vert \mathbf{x}\right\Vert _{1}$,
where $\mathbf{A}\in\mathbb{R}^{N\times K}$ is a known dictionary
and $\mathbf{y}\in\mathbb{R}^{K\times1}$ is the available noisy measurement.
Many algorithms have been proposed for the linear regression problem,
for example, the fast iterative soft-thresholding algorithm (FISTA)
\cite{Beck2009a}, the block coordinate descent (BCD) algorithm \cite{Tseng2001},
the alternating direction method of multiplier (ADMM) \cite{Boyd2010},
proximal algorithm \cite{Parikh2014} and the parallel best-response
with exact line search algorithm \cite{Yang_ConvexApprox}.

In linear regression, the function $f(\mathbf{x})=\frac{1}{2}\left\Vert \mathbf{Ax-y}\right\Vert _{2}^{2}$
is convex in $\mathbf{x}$. This is generally desirable in the design
of numerical algorithms solving problem (\ref{eq:general-formulation})
iteratively. However, this desirable property is not available in
many other applications where we have to deal with a nonconvex $f$.
Consider for example the linear regression model where we assume that
the dictionary $\mathbf{A}$ is unknown and treated as a variable.
In this case, the objective function $f(\mathbf{A},\mathbf{x})=\frac{1}{2}\left\Vert \mathbf{Ax-y}\right\Vert _{2}^{2}$
is a nonconvex function in $(\mathbf{A},\mathbf{x})$ and the problem
is known as Dictionary Learning. In nonlinear regression problems
\cite{Yang2015_arxiv_ICML}, $f(\mathbf{x})$ is in general a nonconvex
function, for example, $f(\mathbf{x})=\frac{1}{2}\left\Vert \boldsymbol{\sigma}\mathbf{(Ax)-b}\right\Vert _{2}^{2}$
and $\boldsymbol{\sigma}$ is a given function specifying the nonlinear
regression model, e.g., the cosine or sigmoid function.

When the function $f$ is nonconvex, the above mentioned algorithms
must be re-examined. For example, the FISTA algorithm no longer converges,
and the generalized iterative soft-thresholding algorithm (GIST) has
been proposed instead. However, as a proximal type algorithm, the
GIST algorithm suffers from slow convergence \cite{Beck2009a}. The
block coordinate descent (BCD) algorithm usually exhibits a faster
convergence because the variable update is is based on the so-called
nonlinear best-response \cite{Bertsekas}: the variable $\mathbf{x}$
is partitioned into multiple block variables $\mathbf{x}=(\mathbf{x}_{k})_{k=1}^{K}$,
and in each iteration of the BCD algorithm, one block variable, say
$\mathbf{x}_{k}$, is updated by its best-response $\mathbf{x}_{k}^{t+1}=\arg\min_{\mathbf{x}_{k}}h(\mathbf{x}_{1}^{t+1},\ldots,\mathbf{x}_{k-1}^{t+1},\mathbf{x}_{k},\mathbf{x}_{k+1}^{t},\ldots,\mathbf{x}_{K}^{t})$
(i.e., the optimal point that minimizes $h(\mathbf{x})$ with respect
to (w.r.t.) the variable $\mathbf{x}_{k}$ only while the remaining
variables are fixed to their values of the preceding iteration) while
all block variables are updated sequentially. Its convergence is guaranteed
under some sufficient conditions on $f$ and $g$ \cite{Tseng2001,Razaviyayn2013,Beck2013,Wright2015},
and due to its simplicity, this method and its variants have been
successfully adopted to many practical problems including the network
anomaly detection problem in \cite{Mardani2013b}. Nevertheless, a
major drawback of the sequential update is that it may incur a large
delay because the $(k+1)$-th block variable $\mathbf{x}_{k+1}$ cannot
be updated until the $k$-th block variable $\mathbf{x}_{k}$ is updated
and the delay may be very large when $K$ is large, which is a norm
rather than an exception in big data analytics \cite{Slavakis2014a}.

A parallel variable update based on the best-response (also known
as the parallel block coordinate descent algorithm \cite{Elad2006})
seems attractive as a mean to speed up the updating procedure, however,
sufficient conditions guaranteeing the convergence of a parallel best-response
algorithm are known for smooth problems only (that is, $g(\mathbf{x})=0$)
and they are rather restrictive, for example, $f$ is convex and satisfies
the diagonal dominance condition \cite{Bertsekas}. However, it has
been shown in some recent works \cite{Elad2006,Razaviyayn2013,Razaviyayn2014,Scutari_BigData}
that if a stepsize is employed in the variable update, the convergence
conditions can be notably relaxed, for example, $f$ could be nonconvex.
Therefore the notion of approximate functions play a fundamental role:
a sequence of successively refined approximate problems are solved,
and the algorithm converges to a stationary point of the original
function $h$ for a number of choices of approximate functions, including
the best-response type approximation, as long as they satisfy some
assumptions on, e.g., (strong or strict) convexity, hence the name
of the successive convex approximation (SCA) framework \cite{Scutari_BigData,Razaviyayn2014}.

The performance of the SCA algorithms in \cite{Scutari_BigData,Razaviyayn2014}
is largely dependent on the choice of the stepsizes, namely, exact/successive
line search and diminishing stepsizes such as constant stepsizes and
diminishing stepsizes. In the (traditional) exact line search (for
example \cite[Sec. III-D]{Elad2006}), a nonconvex nonsmooth optimization
problem must be solved and the complexity is thus high. The successive
line search has a lower complexity, but it typically consists of evaluating
the nonsmooth function $g$ several times for different stepsizes
per iteration \cite[Remark 4]{Scutari_BigData}, which might be computationally
expensive for some $g$ such as the nuclear norm \cite{Steffens2016}.
Diminishing stepsizes has the lowest complexity, but sometimes they
are difficult to deploy in practice because the convergence behavior
is sensitive to the decay rate \cite{Yang_ConvexApprox}. As a matter
of fact, the applicability of SCA algorithms in big data analytics
is severely limited by the meticulous choice of stepsizes \cite{Slavakis2014a}.

To reduce the complexity of the traditional line search schemes and
avoid the parameter tuning of the diminishing stepsize rules, a new
line search scheme is proposed in \cite{Yang_ConvexApprox}: the exact
line search is carried out over a properly constructed differentiable
function; in the successive line search, the approximate function
only needs to be optimized once. The line search schemes in \cite{Yang_ConvexApprox}
are much easier to implement, and closed-form expressions even exist
for many applications. Besides this, the assumption on the strong
or strict convexity of the approximate functions made in \cite{Scutari_BigData,Razaviyayn2014}
is also relaxed to convexity in \cite{Yang_ConvexApprox}.

Another popular algorithm for problem (\ref{eq:general-formulation})
in big data analytics is the alternating direction method of multipliers
(ADMM) \cite{Boyd2010}, but it does not have a guaranteed convergence
to a stationary point if the optimization problem (\ref{eq:general-formulation})
is nonconvex \cite{Mardani2013}. There is some recent development
in ADMM for nonconvex problems, see \cite{Hong2016,Jiang2016} and
the references therein. Nevertheless, the algorithms proposed therein
are for specific problems and not applicable in a broader setup. For
example, the ADMM algorithm proposed in \cite{Hong2016} is designed
for nonconvex sharing/consensus problems, and the ADMM algorithm proposed
in \cite{Jiang2016} converges only when the dictionary matrix has
full row rank, which is generally not satisfied for the network anomaly
detection problem \cite{Mardani2013b}.

So far we have assumed that the regularization function $g$ in (\ref{eq:general-formulation})
is convex, for example, the $\ell_{1}$-norm function, as it has been
used as a standard regularization function to promote sparse solutions
\cite{Tibshirani2011a}. However, it was pointed out in \cite{Fan2001,Candes2008a}
that the $\ell_{1}$-norm is a loose approximation of the $\ell_{0}$-norm
and it tends to produce biased estimates when the sparse signal has
large coefficients. A more desirable regularization function is singular
at the origin while flat elsewhere. Along this direction, several
nonconvex regularization functions have been proposed, for example,
the smoothly clipped absolute deviation \cite{Fan2001}, the capped
$\ell_{1}$-norm \cite{Zhang2010}, and the logarithm function \cite{Weston2003};
we refer the interested reader to \cite{Gasso2009} for a more comprehensive
review.

The nonconvexity of the regularization function $g$ renders many
of the above discussed algorithms inapplicable, including the SCA
framework \cite{Yang_ConvexApprox}, because the nonsmooth function
$g$ is assumed to be convex. It is shown in \cite{Gasso2009} that
if the smooth function $f$ is convex and the nonconvex regularization
function $g$ can be written as the sum of a convex and a concave
function, the classic majorization-minimization (MM) method can be
applied to find a stationary point of (\ref{eq:general-formulation}):
firstly in the majorization step, an upper bound function is obtained
by linearizing the concave regularization function, and then the upper
bound function is minimized in the minimization step; see \cite{Sun2017}
for a recent overview article on the MM algorithms. Nevertheless,
the minimum of the upper bound cannot be expressed by a closed-form
expression and must be found iteratively. The MM method is thus a
two-layer algorithm that involves iterating within iterations and
has a high complexity: a new instance of the upper bound function
is minimized by iterative algorithms at each iteration of the MM method
while minimizing the upper bound functions repeatedly is not a trivial
task, even with a warm start that sets the optimal point of the previous
instance as the initial point of the new instance.

To reduce the complexity of the classic MM method, an upper bound
function based on the proximal type approximation is designed in \cite{Gong2013}
and it is much easier to optimize (see \cite{Attouch2013} for a more
general setup). Although the algorithm converges to a stationary point,
it suffers from two limitations. Firstly, the convergence speed with
the proximal type upper bound functions is usually slower than some
other approximations, for example, the best-response approximation
\cite{Yang_ConvexApprox}. Secondly, the proximal type upper bound
function minimized in each iteration is nonconvex, and it may not
be easy to optimize except in the few cases discussed in \cite{Gong2013}.

In this paper, we propose a SCA framework for problem (\ref{eq:general-formulation})
where the smooth function $f$ is nonconvex and the nonsmooth nonconvex
regularization function $g$ is the difference of two convex functions.\footnote{Some preliminary results of this paper have been presented at \cite{Yang_Rank_ICASSP2018,Yang_MM_SCA_SAM2018}.}
The proposed SCA framework is based on a nontrivial combination of
the SCA framework for a convex $g$ proposed in \cite{Yang_ConvexApprox}
and standard MM framework \cite{Sun2017}. In particular, in each
iteration, we first construct a (possibly nonconvex) \emph{upper bound}
of the original function $h$ by the standard MM method, and then
minimize a \emph{convex approximation} of the upper bound which can
be constructed by the standard SCA framework \cite{Yang_ConvexApprox}.
On the one hand, this is a beneficial combination because the approximate
function is typically much easier to minimize than the original upper
bound function and the proposed algorithm is thus a single layer algorithm
if we choose an approximate function such that its minimum has a closed-form
expression. On the other hand, this is a challenging combination because
the convergence of the proposed algorithms can no longer be proved
by existing techniques. To further speed up the convergence, we design
a line search scheme to calculate the stepsize by generalizing the
line search schemes proposed in \cite{Yang_ConvexApprox} for a convex
$g$. The proposed framework has several attractive features, namely,
\begin{itemize}
\item \emph{flexibility,} as the approximate function does not have to be
a global upper bound of the original objective function and different
choices of the approximate functions lead to different types of algorithms,
for example, proximal type approximation and best-response type approximation;
\item \emph{fast convergence, }as the problem structure can be better exploited
by a proper choice of the approximate function, and the stepsize is
calculated by the line search;
\item \emph{low complexity, }as the approximate function is convex and easy
to optimize, and the proposed line search scheme over a properly constructed
differentiable function is easier to implement than traditional schemes
which are directly applied to the original nonconvex nonsmooth objective
function;
\item \emph{guaranteed convergence to a stationary point,} as long as the
approximate function is convex and satisfies some other mild assumptions
on gradient consistency and continuity.
\end{itemize}
We then illustrate the above attractive features by customizing the
proposed framework for two example applications in subspace learning,
namely, the network anomaly detection problem and the sparse subspace
clustering problem, where both the optimal point of the (best-response
type) approximate functions and the stepsize obtained from the exact
line search have closed-form expressions.

The rest of the paper is organized as follows. In Sec. \ref{sec:Problem-Formulation}
we introduce the problem formulation and the example applications.
The novel SCA framework is proposed and its convergence is analyzed
in Sec. \ref{sec:The-Proposed-SCA}. In Sec. \ref{sec:application-rank-minimization}
and Sec. \ref{sec:application-CappedL1}, two example applications,
the network anomaly detection problem through sparsity regularized
rank minimization and the subspace clustering problem through capped
$\ell_{1}$-norm minimization, are discussed, both theoretically and
numerically. The paper is concluded in Sec. \ref{sec:Concluding-Remarks}.

\emph{Notation: }We use $x$, $\mathbf{x}$ and $\mathbf{X}$ to denote
a scalar, vector and matrix, respectively. We use $X_{jk}$ to denote
the $(j,k)$-th element of $\mathbf{X}$; $x_{k}$ is the $k$-th
element of $\mathbf{x}$ where $\mathbf{x}=(x_{k})_{k=1}^{K}$, and
$\mathbf{x}_{-k}$ denotes all elements of $\mathbf{x}$ except $x_{k}$:
$\mathbf{x}_{-k}=(x_{j})_{j=1,j\neq k}^{K}$. We denote $\mathbf{x}^{-1}$
as the element-wise inverse of $\mathbf{x}$, i.e., $(\mathbf{x}^{-1})_{k}=1/x_{k}$.
Notation $\mathbf{x}\circ\mathbf{y}$ and $\mathbf{X}\otimes\mathbf{Y}$
denotes the Hadamard product between $\mathbf{x}$ and $\mathbf{y}$,
and the Kronecker product between $\mathbf{X}$ and $\mathbf{Y}$,
respectively. The operator $[\mathbf{x}]_{\mathbf{a}}^{\mathbf{b}}$
returns the element-wise projection of $\mathbf{x}$ onto $[\mathbf{a,b}]$:
$[\mathbf{x}]_{\mathbf{a}}^{\mathbf{b}}\triangleq\max(\min(\mathbf{x},\mathbf{b}),\mathbf{a})$.
We denote $\mathbf{d}(\mathbf{X})$ as the vector that consists of
the diagonal elements of $\mathbf{X}$ and $\textrm{diag}(\mathbf{x})$
is a diagonal matrix whose diagonal elements are as same as those
of $\mathbf{x}$. We use $\mathbf{1}$ to denote a vector with all
elements equal to 1. The sign function $\textrm{sign}(x)=1$ if $x>0$,
0 if $x=0$, and $-1$ if $x<0$, and $\textrm{sign}(\mathbf{x})=(\textrm{sign}(x_{k}))_{k}$.

\section{\label{sec:Problem-Formulation}Problem Formulation}

In this section, we formally introduce the problem that will be tackled
in the rest of the paper. In particular, we assume $g(\mathbf{x})$
in (\ref{eq:general-formulation}) can be written as the difference
of two convex functions, and consider from now on the following problem:
\begin{align}
\underset{\mathbf{x}\in\mathcal{X}}{\textrm{minimize}}\quad & h(\mathbf{x})\triangleq f(\mathbf{x})+\underbrace{g^{+}(\mathbf{x})-g^{-}(\mathbf{x})}_{g(\mathbf{x})},\label{eq:problem-formulation}
\end{align}
where
\begin{itemize}
\item $f$ is a proper and differentiable function with a continuous gradient,
\item $g^{+}$ and $g^{-}$ are convex functions, and
\item $\mathcal{X}$ is a closed and convex set.
\end{itemize}
Note that $f(\mathbf{x})$ is not necessarily convex, and $g^{+}(\mathbf{x})$
and $g^{-}(\mathbf{x})$ are not necessarily differentiable.

We aim at developing efficient iterative algorithms that converge
to a stationary point $\mathbf{x}^{\star}$ of problem (\ref{eq:problem-formulation})
that satisfies the first order optimality condition:
\[
(\mathbf{x}-\mathbf{x}^{\star})^{T}(\nabla f(\mathbf{x}^{\star})+\boldsymbol{\xi}^{+}(\mathbf{x}^{\star})-\boldsymbol{\xi}^{-}(\mathbf{x}^{\star}))\geq0,\forall\mathbf{x}\in\mathcal{X},
\]
where $\boldsymbol{\xi}^{+}(\mathbf{x})$ and $\boldsymbol{\xi}^{-}(\mathbf{x})$
is a subgradient of $g^{+}(\mathbf{x})$ and $g^{-}(\mathbf{x})$,
respectively. Note that a convex function always has a subgradient.

\subsection{Example Application: Network Anomaly Detection Through Sparsity Regularized
Rank Minimization}

Consider the problem of estimating a low rank matrix $\mathbf{X}\in\mathbb{R}^{N\times K}$
and a sparse matrix $\mathbf{S}\in\mathbb{R}^{I\times K}$ from the
noisy measurement $\mathbf{Y}\in\mathbb{R}^{N\times K}$ which is
the output of a linear system:
\[
\mathbf{Y}=\mathbf{X}+\mathbf{DS}+\mathbf{V},
\]
where $\mathbf{D}\in\mathbb{R}^{N\times I}$ is known and $\mathbf{V}^{N\times K}$
is the unknown noise.

The rank of $\mathbf{X}$ is much smaller than $N$ and $K$, i.e,
$\textrm{rank}(\mathbf{X})\ll\min(N,K)$, and the support size of
$\mathbf{S}$ is much smaller than $IK$, i.e., $\left\Vert \mathbf{S}\right\Vert _{0}\ll IK$.
A natural measure for the estimation error is the least square loss
function augmented by regularization functions to promote the rank
sparsity of $\mathbf{X}$ and support sparsity of $\mathbf{S}$:
\begin{align}
\underset{\mathbf{X},\mathbf{S}}{\textrm{minimize}}\quad & \frac{1}{2}\left\Vert \mathbf{X}+\mathbf{D}\mathbf{S}-\mathbf{Y}\right\Vert _{F}^{2}+\lambda\left\Vert \mathbf{X}\right\Vert _{*}+\mu\left\Vert \mathbf{S}\right\Vert _{1},\label{eq:SRRM}
\end{align}
where $\left\Vert \mathbf{X}\right\Vert _{*}$ is the nuclear norm
of $\mathbf{X}$. Problem (\ref{eq:SRRM}) plays a fundamental role
in the analysis of traffic anomalies in large-scale backbone networks
\cite{Mardani2013b}. In this application, $\mathbf{D}$ is a given
binary routing matrix, $\mathbf{X}=\mathbf{R}\mathbf{Z}$ where $\mathbf{Z}$
is the unknown traffic flows over the time horizon of interest, and
$\mathbf{S}$ is the traffic volume anomalies. The matrix $\mathbf{X}$
inherits the rank sparsity from $\mathbf{Z}$ because common temporal
patterns among the traffic flows in addition to their periodic behavior
render most rows/columns of $\mathbf{Z}$ linearly dependent and thus
low rank, and $\mathbf{S}$ is assumed to be sparse because traffic
anomalies are expected to happen sporadically and last shortly relative
to the measurement interval, which is represented by the number of
columns $K$.

Problem (\ref{eq:SRRM}) is convex and it can be solved by the SCA
algorithm proposed in \cite{Steffens2016}, which is a parallel best-response
with exact line search algorithm. Although it presents a much lower
complexity than standard methods such as proximal type algorithms
and BCD algorithms, it may eventually become inefficient due to the
use of complex models: computing the nuclear norm $\left\Vert \mathbf{X}\right\Vert _{*}$
has a cubic complexity and is unaffordable when the problem dimension
is large. Furthermore, problem (\ref{eq:SRRM}) is not suitable for
the design of distributed and/or parallel algorithms because the nuclear
norm $\left\Vert \mathbf{X}\right\Vert _{*}$ is neither differentiable
nor decomposable among the blocks of $\mathbf{X}$ (unless $\mathbf{X}$
is Hermitian).

It follows from the identity \cite{Burer2003,Recht2010}
\[
\left\Vert \mathbf{X}\right\Vert _{*}=\min_{(\mathbf{P},\mathbf{Q})}\frac{1}{2}\left(\left\Vert \mathbf{P}\right\Vert _{F}^{2}+\left\Vert \mathbf{Q}\right\Vert _{F}^{2}\right),\textrm{ s.t. }\mathbf{P}\mathbf{Q}=\mathbf{X}
\]
that the low rank matrix $\mathbf{X}$ can be written according to
the above matrix factorization as the product of two low rank matrices
$\mathbf{P}\in\mathbb{R}^{N\times\rho}$ and $\mathbf{Q}\in\mathbb{R}^{\rho\times K}$
for a $\rho$ that is larger than the rank of $\mathbf{X}$ but usually
much smaller than $N$ and $K$: $\textrm{rank}(\mathbf{X})\leq\rho\ll\min(N,K)$.
It may be useful to consider the following optimization problem where
the nuclear norm $\left\Vert \mathbf{X}\right\Vert _{*}$ is replaced
by $\left\Vert \mathbf{P}\right\Vert _{F}^{2}+\left\Vert \mathbf{Q}\right\Vert _{F}^{2}$,
which is differentiable and separable among its blocks:
\begin{equation}
\underset{\mathbf{P},\mathbf{Q},\mathbf{S}}{\textrm{minimize}}\;\frac{1}{2}\left\Vert \mathbf{P}\mathbf{Q}+\mathbf{D}\mathbf{S}-\mathbf{Y}\right\Vert _{F}^{2}+\frac{\lambda}{2}\left(\left\Vert \mathbf{P}\right\Vert _{F}^{2}+\left\Vert \mathbf{Q}\right\Vert _{F}^{2}\right)+\mu\left\Vert \mathbf{S}\right\Vert _{1}.\label{eq:eq:rank-problem-formulation-copy}
\end{equation}
This optimization problem is a special case of (\ref{eq:problem-formulation})
obtained by setting
\begin{align*}
f(\mathbf{P},\mathbf{Q},\mathbf{S}) & \triangleq\frac{1}{2}\left\Vert \mathbf{P}\mathbf{Q}+\mathbf{D}\mathbf{S}-\mathbf{Y}\right\Vert _{F}^{2}+\frac{\lambda}{2}\left(\left\Vert \mathbf{P}\right\Vert _{F}^{2}+\left\Vert \mathbf{Q}\right\Vert _{F}^{2}\right),\\
g^{+}(\mathbf{S}) & \triangleq\mu\left\Vert \mathbf{S}\right\Vert _{1},\textrm{ and }g^{-}(\mathbf{S})=0.
\end{align*}
Although problem (\ref{eq:eq:rank-problem-formulation-copy}) is nonconvex,
every stationary point of (\ref{eq:eq:rank-problem-formulation-copy})
is an optimal solution of (\ref{eq:SRRM}) under some mild conditions
\cite[Prop. 1]{Mardani2013}. In Sec. \ref{sec:application-rank-minimization},
we will customize the proposed SCA framework to design an iterative
soft-thresholding with exact line search algorithm for problem (\ref{eq:eq:rank-problem-formulation-copy}),
which is essentially a parallel best-response algorithm.

\subsection{Example Application: Sparse Subspace Clustering Through Capped $\ell_{1}$-Norm
Minimization}

\begin{figure}[t]
\center

\includegraphics[scale=0.5]{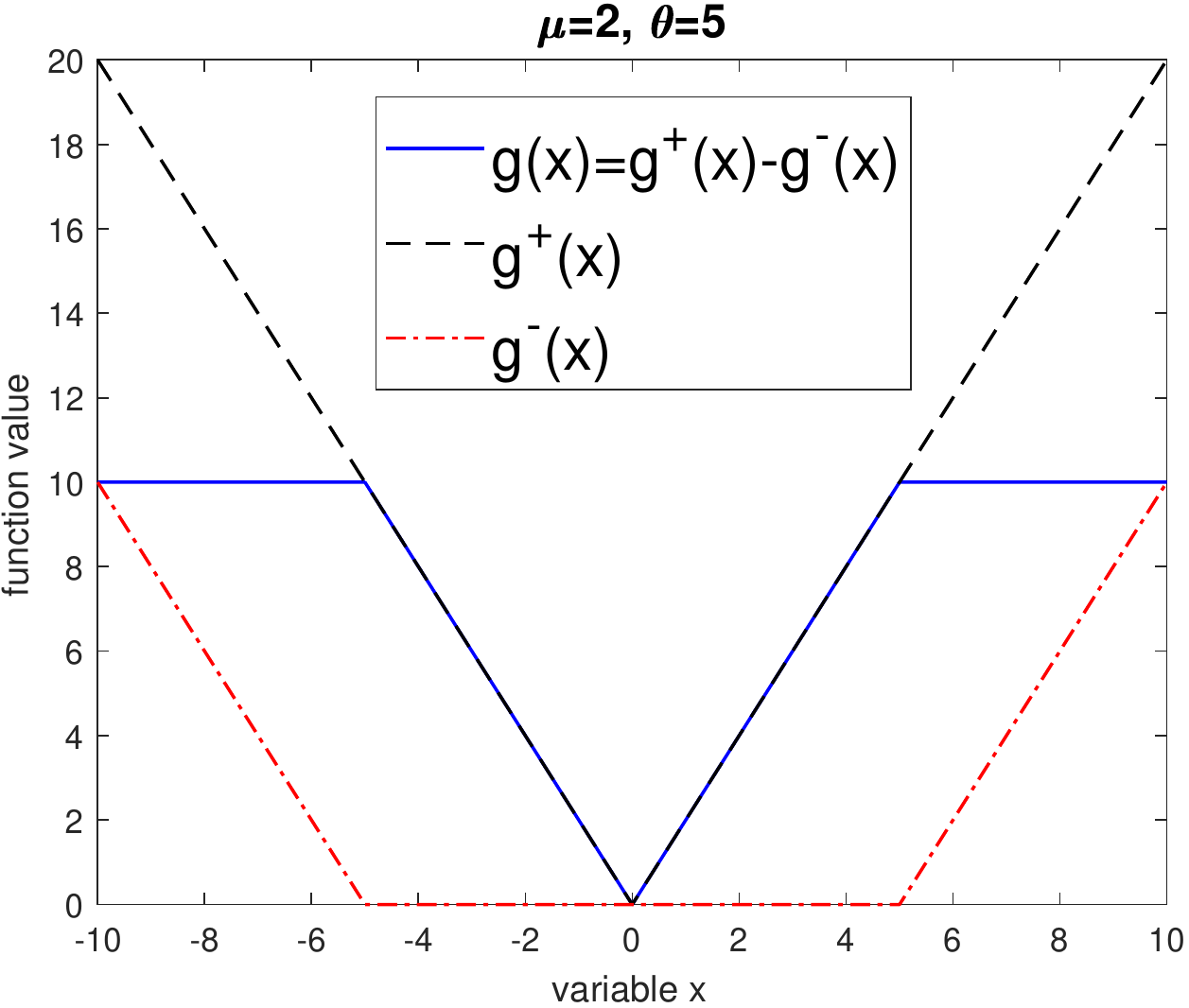}\caption{\label{fig:Illustration-of-Decomposition}Illustration of the capped
$\ell_{1}$-norm function and its decomposition}
\end{figure}

Consider the linear regression model
\[
\mathbf{y=Ax+v},
\]
where the dictionary $\mathbf{A}\in\mathbb{R}^{N\times K}$ is known
and $\mathbf{y}\in\mathbb{R}^{N\times1}$ is the noisy measurement.
To estimate $\mathbf{x}$ which is known to be sparse a priori, we
minimize the quadratic estimation error function augmented by some
regularization function to promote the sparsity of $\mathbf{x}$.
A common routine is to use the $\ell_{1}$-norm, which has however
been shown to yield biased estimates for large coefficients \cite{Zhang2010}.
Alternatives include for example the capped $\ell_{1}$-norm function
\cite{Zhang2010,Gasso2009,Gong2013}, and the resulting optimization
problem is as follows:
\begin{equation}
\underset{\mathbf{x}}{\textrm{minimize}}\quad\frac{1}{2}\left\Vert \mathbf{Ax-y}\right\Vert _{2}^{2}+\mu\sum_{k=1}^{K}\min(|x_{k}|,\theta).\label{eq:cappedL1-scalar-formulation}
\end{equation}
This optimization problem is a special case of (\ref{eq:problem-formulation})
obtained by setting
\begin{align*}
f(\mathbf{x}) & \triangleq\frac{1}{2}\left\Vert \mathbf{Ax-b}\right\Vert _{2}^{2},\\
g^{+}(\mathbf{x}) & \triangleq\mu\sum_{k=1}^{K}|x_{k}|,\textrm{ and }g^{-}(\mathbf{x})\triangleq\mu\sum_{k=1}^{K}|x_{k}|-\min(|x_{k}|,\theta),
\end{align*}
where $g^{-}(\mathbf{x})$ is a convex but nonsmooth function. A graphical
illustration of the functions $g$, $g^{+}$ and $g^{-}$ is provided
in Fig. \ref{fig:Illustration-of-Decomposition}, and interested readers
are referred to \cite[Fig. 2]{Gasso2009} for more examples.

When $\theta$ is sufficiently large, problem (\ref{eq:cappedL1-scalar-formulation})
reduces to the standard LASSO problem, which plays a fundamental role
in sparse subspace clustering problems \cite{Elhamifar2013} and can
be solved efficiently by the SCA algorithm proposed in \cite{Yang_ConvexApprox}.
In Problem (\ref{eq:cappedL1-scalar-formulation}), we take one step
further by considering the capped $\ell_{1}$-norm and then in Sec.
\ref{sec:application-CappedL1}, we customize the proposed SCA framework
to design an iterative soft-thresholding with exact line search algorithm
for problem (\ref{eq:cappedL1-scalar-formulation}), which is essentially
a parallel best-response algorithm.

\section{\label{sec:The-Proposed-SCA}The Proposed Successive Convex Approximation
Algorithms}

In this section, we formally introduce the proposed SCA framework
for problem (\ref{eq:problem-formulation}), where $h$ is in general
a nonconvex function since $f$ is not necessarily convex and $g^{-}$
is convex, and $h$ is in general a nonsmooth function since both
$g^{+}$ and $g^{-}$ are assumed to be nonsmooth.

At any arbitrary but given point $\mathbf{x}^{t}$, assume the subgradient
of $g^{-}(\mathbf{x})$ is $\boldsymbol{\xi}^{-}(\mathbf{x}^{t})$.
Since $g^{-}$ is convex, it follows from Jensen's inequality that
\begin{equation}
g^{-}(\mathbf{x})\geq g^{-}(\mathbf{x}^{t})+(\mathbf{x}-\mathbf{x}^{t})^{T}\boldsymbol{\xi}^{-}(\mathbf{x}^{t}),\forall\mathbf{x}\in\mathcal{X}.\label{eq:definition-of-convex-functions}
\end{equation}
Define $\overline{h}(\mathbf{x};\mathbf{x}^{t})$ as
\begin{equation}
\overline{h}(\mathbf{x};\mathbf{x}^{t})\triangleq f(\mathbf{x})-g^{-}(\mathbf{x}^{t})-(\mathbf{x}-\mathbf{x}^{t})^{T}\boldsymbol{\xi}^{-}(\mathbf{x}^{t})+g^{+}(\mathbf{x}).\label{eq:upper-bound-1}
\end{equation}
We can readily infer from (\ref{eq:definition-of-convex-functions})
that $\overline{h}(\mathbf{x};\mathbf{x}^{t})$ is a global upper
bound of $h(\mathbf{x})$ which is tight at $\mathbf{x}=\mathbf{x}^{t}$:
\begin{equation}
\overline{h}(\mathbf{x};\mathbf{x}^{t})\geq h(\mathbf{x}),\textrm{ and }\overline{h}(\mathbf{x}^{t};\mathbf{x}^{t})=h(\mathbf{x}^{t}),\forall\mathbf{x}\in\mathcal{X}.\label{eq:upper-bound-2}
\end{equation}
In the standard MM method for problem (\ref{eq:problem-formulation})
proposed in \cite{Gasso2009}, a sequence of points $\{\mathbf{x}^{t}\}_{t}$
is generated by minimizing the upper bound function $\overline{h}(\mathbf{x};\mathbf{x}^{t})$:
\begin{equation}
\mathbf{x}^{t+1}=\underset{\mathbf{x}\in\mathcal{X}}{\arg\min}\;\overline{h}(\mathbf{x};\mathbf{x}^{t}).\label{eq:CCCP}
\end{equation}
This and (\ref{eq:upper-bound-2}) imply that $\{h(\mathbf{x}^{t})\}_{t}$
is a decreasing sequence as
\[
h(\mathbf{x}^{t+1})\leq\overline{h}(\mathbf{x}^{t+1};\mathbf{x}^{t})\leq\overline{h}(\mathbf{x}^{t};\mathbf{x}^{t})=h(\mathbf{x}^{t}).
\]
However, the optimization problem (\ref{eq:CCCP}) is not necessarily
easy to solve due to two possible reasons: $\overline{h}(\mathbf{x};\mathbf{x}^{t})$
may be nonconvex, and $\mathbf{x}^{t+1}$ may not have a closed-form
expression and must be found iteratively.

The proposed algorithm consists of minimizing a sequence of successively
refined approximate functions. Given $\mathbf{x}^{t}$ at iteration
$t$, we propose to minimize a properly designed \emph{approximate
function} of the \emph{upper bound function} $\overline{h}(\mathbf{x};\mathbf{x}^{t})$,
denoted as $\tilde{h}(\mathbf{x};\mathbf{x}^{t})$:

\begin{equation}
\tilde{h}(\mathbf{x};\mathbf{x}^{t})=\tilde{f}(\mathbf{x};\mathbf{x}^{t})-(\mathbf{x}-\mathbf{x}^{t})^{T}\boldsymbol{\xi}^{-}(\mathbf{x}^{t})+g^{+}(\mathbf{x}),\label{eq:approximate-function}
\end{equation}
where $\tilde{f}(\mathbf{x};\mathbf{x}^{t})$ is an approximate function
of $f(\mathbf{x})$ at $\mathbf{x}^{t}$ that satisfies several technical
conditions that are in the same essence as those specified in \cite{Yang_ConvexApprox},
namely,

\noindent (A1) The approximate function $\tilde{f}(\mathbf{x};\mathbf{x}^{t})$
is convex in $\mathbf{x}$ for any given $\mathbf{x}^{t}\in\mathcal{X}$;

\noindent (A2) The approximate function $\tilde{f}(\mathbf{x};\mathbf{x}^{t})$
is continuously differentiable in $\mathbf{x}$ for any given $\mathbf{x}^{t}\in\mathcal{X}$
and continuous in $\mathbf{x}^{t}$ for any $\mathbf{x}\in\mathcal{X}$;

\noindent (A3) The gradient of $\tilde{f}(\mathbf{x};\mathbf{x}^{t})$
and the gradient of $f(\mathbf{x})$ are identical at $\mathbf{x}=\mathbf{x}^{t}$
for any $\mathbf{x}^{t}\in\mathcal{X}$, i.e., $\nabla_{\mathbf{x}}\tilde{f}(\mathbf{x}^{t};\mathbf{x}^{t})=\nabla_{\mathbf{x}}f(\mathbf{x}^{t})$.

\noindent Comparing $\overline{h}(\mathbf{x};\mathbf{x}^{t})$ in
(\ref{eq:upper-bound-1}) with $\tilde{h}(\mathbf{x};\mathbf{x}^{t})$
in (\ref{eq:approximate-function}), we see that replacing $f(\mathbf{x})$
in $\overline{h}(\mathbf{x};\mathbf{x}^{t})$ by its approximate function
$\tilde{f}(\mathbf{x};\mathbf{x}^{t})$ leads to the proposed approximate
function $\tilde{h}(\mathbf{x};\mathbf{x}^{t})$. Note that $\tilde{h}(\mathbf{x};\mathbf{x}^{t})$
is not necessarily a global upper bound of $\overline{h}(\mathbf{x};\mathbf{x}^{t})$
(or the original function $h(\mathbf{x})$), because according to
Assumptions (A1)-(A3), $\tilde{f}(\mathbf{x};\mathbf{x}^{t})$ does
not have to be a global upper bound of $f(\mathbf{x})$.

At iteration $t$, the approximate problem consists of minimizing
the approximate function $\tilde{h}(\mathbf{x};\mathbf{x}^{t})$ over
the same constraint set $\mathcal{X}$:
\begin{align}
\underset{\mathbf{x}\in\mathcal{X}}{\textrm{minimize}}\; & \underbrace{\tilde{f}(\mathbf{x};\mathbf{x}^{t})-(\mathbf{x}-\mathbf{x}^{t})^{T}\boldsymbol{\xi}^{-}(\mathbf{x}^{t})+g^{+}(\mathbf{x})}_{\tilde{h}(\mathbf{x};\mathbf{x}^{t})}.\label{eq:approximate-problem}
\end{align}
Since $\tilde{f}(\mathbf{x};\mathbf{x}^{t})$ is convex by assumption
(A1), (\ref{eq:approximate-problem}) is a convex optimization problem.
We denote as $\mathbb{B}\mathbf{x}^{t}$ an (globally) optimal solution
of (\ref{eq:approximate-problem}) and as $\mathcal{S}(\mathbf{x}^{t})$
the set of (globally) optimal solutions:
\begin{equation}
\mathbb{B}\mathbf{x}^{t}\in\mathcal{S}(\mathbf{x}^{t})=\left\{ \mathbf{x}^{\star}:\mathbf{x}^{\star}\in\underset{\mathbf{x}\in\mathcal{X}}{\arg\min}\;\tilde{h}(\mathbf{x};\mathbf{x}^{t})\right\} .\label{eq:best-response}
\end{equation}
Based on (\ref{eq:best-response}), we define the mapping $\mathbb{B}\mathbf{x}$
that is used to generate the sequence of points in the proposed algorithm:
\begin{equation}
\mathcal{X}\ni\mathbf{x}\longmapsto\mathbb{B}\mathbf{x}\in\mathcal{X}.\label{eq:mapping}
\end{equation}
Given the mapping $\mathbb{B}\mathbf{x}$, the following properties
hold.
\begin{prop}
[Stationary point and descent direction]\label{prop:descent-property}
Provided that Assumptions (A1)-(A3) are satisfied: (i) A point $\mathbf{x}^{t}$
is a stationary point of (\ref{eq:problem-formulation}) if and only
if $\mathbf{x}^{t}\in\mathcal{S}(\mathbf{x}^{t})$ defined in (\ref{eq:best-response});
(ii) If $\mathbf{x}^{t}$ is not a stationary point of (\ref{eq:best-response}),
then $\mathbb{B}\mathbf{x}^{t}-\mathbf{x}^{t}$ is a descent direction
of $\overline{h}(\mathbf{x};\mathbf{x}^{t})$ at $\mathbf{x}=\mathbf{x}^{t}$
in the sense that
\begin{equation}
(\mathbb{B}\mathbf{x}^{t}-\mathbf{x}^{t})^{T}(\nabla f(\mathbf{x}^{t})-\boldsymbol{\xi}^{-}(\mathbf{x}^{t}))+g^{+}(\mathbb{B}\mathbf{x}^{t})-g^{+}(\mathbf{x}^{t})<0.\label{eq:descent-direction}
\end{equation}
\end{prop}
\begin{IEEEproof}
See Appendix \ref{sec:Proof-of-Propositions}.
\end{IEEEproof}
 \newcounter{MYtempeqncnt} \begin{figure*}[t] \normalsize \setcounter{MYtempeqncnt}{\value{equation}} \setcounter{equation}{20} \vspace*{4pt}
\begin{align}
 & f(\mathbf{x}^{t}+\beta^{m}(\mathbb{B}\mathbf{x}^{t}-\mathbf{x}^{t}))-(\mathbf{x}^{t}+\beta^{m}(\mathbb{B}\mathbf{x}^{t}-\mathbf{x}^{t})-\mathbf{x}^{t})^{T}\boldsymbol{\xi}^{-}(\mathbf{x}^{t})+g^{+}(\mathbf{x}^{t})+\beta^{m}(g^{+}(\mathbb{B}\mathbf{x}^{t})-g^{+}(\mathbf{x}^{t}))\nonumber \\
\leq\; & f(\mathbf{x}^{t})+g^{+}(\mathbf{x}^{t})+\alpha\beta^{m}((\mathbb{B}\mathbf{x}^{t}-\mathbf{x}^{t})^{T}(\nabla f(\mathbf{x}^{t})-\boldsymbol{\xi}^{-}(\mathbf{x}^{t}))+g^{+}(\mathbb{B}\mathbf{x}^{t})-g^{+}(\mathbf{x}^{t})).\label{eq:successive-line-search-proposed}\\
 & f(\mathbf{x}^{t}+\beta^{m}(\mathbb{B}\mathbf{x}^{t}-\mathbf{x}^{t}))-\beta^{m}(\mathbb{B}\mathbf{x}^{t}-\mathbf{x}^{t})^{T}\boldsymbol{\xi}^{-}(\mathbf{x}^{t})+\beta^{m}(g^{+}(\mathbb{B}\mathbf{x}^{t})-g^{+}(\mathbf{x}^{t}))\nonumber \\
\leq\; & f(\mathbf{x}^{t})+\alpha\beta^{m}((\mathbb{B}\mathbf{x}^{t}-\mathbf{x}^{t})^{T}(\nabla f(\mathbf{x}^{t})-\boldsymbol{\xi}^{-}(\mathbf{x}^{t}))+g^{+}(\mathbb{B}\mathbf{x}^{t})-g^{+}(\mathbf{x}^{t})).\label{eq:successive-line-search-proposed-simplified}
\end{align}
\setcounter{equation}{\value{MYtempeqncnt}} \hrulefill  \end{figure*}

If $\mathbb{B}\mathbf{x}^{t}-\mathbf{x}^{t}$ is a descent direction
of $\overline{h}(\mathbf{x};\mathbf{x}^{t})$ at $\mathbf{x}=\mathbf{x}^{t}$,
there exists a scalar $\gamma^{t}\in(0,1]$ such that
\[
\overline{h}(\mathbf{x}^{t}+\gamma^{t}(\mathbb{B}\mathbf{x}^{t}-\mathbf{x}^{t}))<\overline{h}(\mathbf{x}^{t}),
\]
for which a formal proof is provided shortly in Proposition \ref{prop:stepsize}.
This motivates us to update the variable as follows
\begin{equation}
\mathbf{x}^{t+1}=\mathbf{x}^{t}+\gamma^{t}(\mathbb{B}\mathbf{x}^{t}-\mathbf{x}^{t}).\label{eq:variable-update}
\end{equation}
The function value $h(\mathbf{x}^{t})$ is monotonically decreasing
because
\begin{equation}
h(\mathbf{x}^{t+1})\overset{(a)}{\leq}\overline{h}(\mathbf{x}^{t+1};\mathbf{x}^{t})<\overline{h}(\mathbf{x}^{t};\mathbf{x}^{t})\overset{(b)}{=}h(\mathbf{x}^{t}).\label{eq:value-decrease}
\end{equation}
where (\emph{a}) and (\emph{b}) in (\ref{eq:value-decrease}) follow
from (\ref{eq:upper-bound-2}).

There are several commonly used stepsize rules, for example, the constant/decreasing
stepsize rules and the line search. In this paper, we restrict the
discussion to the line search schemes because they lead to a fast
convergence speed as shown in \cite{Yang_ConvexApprox}. On the one
hand, the traditional exact line search aims at finding the optimal
stepsize, denoted as $\gamma_{\textrm{opt}}^{t}$ (\textquotedbl{}opt\textquotedbl{}
stands for \textquotedbl{}optimal\textquotedbl{}) that yields the
largest decrease of $h(\mathbf{x})$ along the direction $\mathbb{B}\mathbf{x}^{t}-\mathbf{x}^{t}$
\cite{Elad2006}: \begin{subequations}\label{eq:exact-line-search-traditional}
\begin{align}
\gamma_{\textrm{opt}}^{t} & \triangleq\underset{0\leq\gamma\leq1}{\arg\min}\;h(\mathbf{x}^{t}+\gamma(\mathbb{B}\mathbf{x}^{t}-\mathbf{x}^{t}))\nonumber \\
 & =\underset{0\leq\gamma\leq1}{\arg\min}\;\left\{ \begin{array}{l}
f(\mathbf{x}^{t}+\gamma(\mathbb{B}\mathbf{x}^{t}-\mathbf{x}^{t}))\smallskip\\
+g^{+}(\mathbf{x}^{t}+\gamma(\mathbb{B}\mathbf{x}^{t}-\mathbf{x}^{t}))\smallskip\\
-g^{-}(\mathbf{x}^{t}+\gamma(\mathbb{B}\mathbf{x}^{t}-\mathbf{x}^{t}))
\end{array}\right\} .\label{eq:exact-line-search-traditional-original}
\end{align}
Although it is a scalar problem, it is not necessarily easy to solve
because it is nonconvex (even when $f(\mathbf{x})$ is convex) and
nondifferentiable. On the other hand, as $\mathbb{B}\mathbf{x}^{t}-\mathbf{x}^{t}$
is also a descent direction of $\overline{h}(\mathbf{x};\mathbf{x}^{t})$
according to Proposition \ref{prop:descent-property}, it is possible
to perform the exact line search over the upper bound function $\overline{h}(\mathbf{x};\mathbf{x}^{t})$
along the direction $\mathbb{B}\mathbf{x}^{t}-\mathbf{x}^{t}$:
\begin{align}
\gamma_{\textrm{ub}}^{t} & \triangleq\underset{0\leq\gamma\leq1}{\arg\min}\;\overline{h}(\mathbf{x}^{t}+\gamma(\mathbb{B}\mathbf{x}^{t}-\mathbf{x}^{t});\mathbf{x}^{t})\nonumber \\
 & =\underset{0\leq\gamma\leq1}{\arg\min}\;\left\{ \begin{array}{l}
f(\mathbf{x}^{t}+\gamma(\mathbb{B}\mathbf{x}^{t}-\mathbf{x}^{t}))\smallskip\\
-(\mathbf{x}^{t}+\gamma(\mathbb{B}\mathbf{x}^{t}-\mathbf{x}^{t})-\mathbf{x}^{t})^{T}\boldsymbol{\xi}^{-}(\mathbf{x}^{t})\smallskip\\
g^{+}(\mathbf{x}^{t}+\gamma(\mathbb{B}\mathbf{x}^{t}-\mathbf{x}^{t}))
\end{array}\right\} ,\label{eq:exact-line-search-traditional-upper-bound}
\end{align}
\end{subequations}and we denote as $\gamma_{\textrm{ub}}^{t}$ (\textquotedbl{}ub\textquotedbl{}
stands for \textquotedbl{}upper bound\textquotedbl{}) the obtained
stepsize. However, this is not always favorable in practice either
because the above minimization problem involves the nonsmooth function
$g^{+}$.

To reduce the complexity of traditional exact line search schemes
in (\ref{eq:exact-line-search-traditional}), we start from (\ref{eq:exact-line-search-traditional-upper-bound}):
applying the Jensen's inequality to the convex function $g^{+}$ in
(\ref{eq:exact-line-search-traditional-upper-bound}) yields that
for any $\gamma\in[0,1]$,
\begin{align}
g^{+}(\mathbf{x}^{t}+\gamma(\mathbb{B}\mathbf{x}^{t}-\mathbf{x}^{t})) & \leq(1-\gamma^{t})g^{+}(\mathbf{x}^{t})+\gamma g^{+}(\mathbb{B}\mathbf{x}^{t})\nonumber \\
 & =g^{+}(\mathbf{x}^{t})+\gamma(g^{+}(\mathbb{B}\mathbf{x}^{t})-g^{+}(\mathbf{x}^{t})).\label{eq:Jensen's inequality}
\end{align}
The function on the right hand side of (\ref{eq:Jensen's inequality})
is a differentiable and linear function in $\gamma$. We thus propose
to perform the line search over the following function which is obtained
by replacing the nonsmooth function $g^{+}$ in (\ref{eq:exact-line-search-traditional-upper-bound})
by its upper bound (\ref{eq:Jensen's inequality}):
\begin{align}
\gamma^{t} & =\underset{0\leq\gamma\leq1}{\arg\min}\left\{ \begin{array}{l}
f(\mathbf{x}^{t}+\gamma(\mathbb{B}\mathbf{x}^{t}-\mathbf{x}^{t}))\smallskip\\
-(\mathbf{x}^{t}+\gamma(\mathbb{B}\mathbf{x}^{t}-\mathbf{x}^{t})-\mathbf{x}^{t})^{T}\boldsymbol{\xi}^{-}(\mathbf{x}^{t})\smallskip\\
+g(\mathbf{x}^{t})+\gamma(g^{+}(\mathbb{B}\mathbf{x}^{t})-g^{+}(\mathbf{x}^{t})).
\end{array}\right\} \nonumber \\
 & =\underset{0\leq\gamma\leq1}{\arg\min}\left\{ \negthickspace\negthickspace\begin{array}{l}
f(\mathbf{x}^{t}+\gamma(\mathbb{B}\mathbf{x}^{t}-\mathbf{x}^{t}))\smallskip\\
+\gamma(g^{+}(\mathbb{B}\mathbf{x}^{t})-g^{+}(\mathbf{x}^{t})-(\mathbb{B}\mathbf{x}^{t}-\mathbf{x}^{t})^{T}\boldsymbol{\xi}^{-}(\mathbf{x}^{t}))
\end{array}\negthickspace\negthickspace\right\} .\label{eq:exact-line-search-proposed}
\end{align}
Combining (\ref{eq:exact-line-search-traditional-upper-bound}) and
(\ref{eq:Jensen's inequality}), we readily see that the function
in (\ref{eq:exact-line-search-traditional-upper-bound}) is upper
bounded by the function in (\ref{eq:exact-line-search-proposed})
which is tight at $\gamma=0$. The optimization problem in (\ref{eq:exact-line-search-proposed})
is differentiable and presumably much easier to optimize than the
nondifferentiable problems in (\ref{eq:exact-line-search-traditional}).
It is furthermore convex if $f(\mathbf{x})$ is convex, and it can
be solved efficiently by the bisection method; in many cases closed-form
expressions even exist, as we will show later by the example applications
in Sec. \ref{sec:application-rank-minimization}-\ref{sec:application-CappedL1}.
This is a desirable property because the scalar optimization problem
in (\ref{eq:exact-line-search-proposed}) is convex as long as $f$
is convex, although the original function $h$ is still not convex
due to $g^{-}$.

Albeit the low complexity, a natural question to ask is whether the
stepsize $\gamma^{t}$ obtained by the proposed exact line search
scheme (\ref{eq:exact-line-search-proposed}) leads to a strict decrease
of the original objective function $h(\mathbf{x})$.\footnote{With a slight abuse of terminology, we call the proposed line search
scheme (\ref{eq:exact-line-search-proposed}) the exact line search,
although it is carried out over a differentiable upper bound of the
original objective function $h$.} The answer is affirmative and we first provide an intuitive explanation:
the gradient of the function in (\ref{eq:exact-line-search-proposed})
w.r.t. $\gamma$ at $\gamma=0$ is
\[
(\mathbb{B}\mathbf{x}^{t}-\mathbf{x}^{t})^{T}\nabla f(\mathbf{x}^{t})+g^{+}(\mathbb{B}\mathbf{x}^{t})-g^{+}(\mathbf{x}^{t})-(\mathbb{B}\mathbf{x}^{t}-\mathbf{x}^{t})^{T}\boldsymbol{\xi}^{-}(\mathbf{x}^{t}),
\]
which is strictly smaller than 0 according to Proposition \ref{prop:descent-property}.
This implies the function has a negative slope at $\gamma=0$ and
its minimum point $\gamma^{t}$ is thus nonzero and positive. Consequently
the objective function $h(\mathbf{x})$ can be strictly decreased:
$h(\mathbf{x}^{t}+\gamma^{t}(\mathbb{B}\mathbf{x}^{t}-\mathbf{x}^{t}))<h(\mathbf{x}^{t})$.
This intuitive explanation will be made rigorous shortly in Proposition
\ref{prop:stepsize}.

If no structure in $f(\mathbf{x})$ (e.g., convexity) can be exploited
to efficiently compute $\gamma^{t}$ according to the exact line search
(\ref{eq:exact-line-search-proposed}), we adopt a stepsize if it
yields sufficient decrease in the sense specified by the successive
line search (also known as the Armijo rule) \cite{bertsekas1999nonlinear}:
given scalars $0<\alpha<1$ and $0<\beta<1$, the stepsize $\gamma^{t}$
is set to be $\gamma^{t}=\beta^{m_{t}}$, where $m_{t}$ is the smallest
nonnegative integer $m$ satisfying the following inequality:\begin{subequations}\label{eq:successive-line-search-traditional}
\begin{align}
h & (\mathbf{x}^{t}+\beta^{m}(\mathbb{B}\mathbf{x}^{t}-\mathbf{x}^{t}))-h(\mathbf{x}^{t})\nonumber \\
 & \leq\alpha\beta^{m}((\mathbb{B}\mathbf{x}^{t}-\mathbf{x}^{t})^{T}(\nabla f(\mathbf{x}^{t})-\boldsymbol{\xi}^{-}(\mathbf{x}^{t}))+g^{+}(\mathbb{B}\mathbf{x}^{t})-g^{+}(\mathbf{x}^{t})),\label{eq:successive-line-search-traditional-original}
\end{align}
or
\begin{align}
\overline{h} & (\mathbf{x}^{t}+\beta^{m}(\mathbb{B}\mathbf{x}^{t}-\mathbf{x}^{t}))-\overline{h}(\mathbf{x}^{t})\nonumber \\
 & \leq\alpha\beta^{m}((\mathbb{B}\mathbf{x}^{t}-\mathbf{x}^{t})^{T}(\nabla f(\mathbf{x}^{t})-\boldsymbol{\xi}^{-}(\mathbf{x}^{t}))+g^{+}(\mathbb{B}\mathbf{x}^{t})-g^{+}(\mathbf{x}^{t})).\label{eq:successive-line-search-traditional-upper-bound}
\end{align}
\end{subequations}As a result, $g^{+}(\mathbf{x}^{t}+\beta^{m}(\mathbb{B}\mathbf{x}^{t}-\mathbf{x}^{t}))$
(in $h$ or $\overline{h}$) must be evaluated for $m_{t}+1$ times,
namely, $m=0,1,\ldots,m_{t}$, and this may incur a high complexity,
for example, when $g^{+}$ is the nuclear norm.

To reduce the complexity of traditional successive line search schemes
(\ref{eq:successive-line-search-traditional}), we follow the reasoning
from (\ref{eq:exact-line-search-traditional}) to (\ref{eq:exact-line-search-proposed})
and propose a successive line search that works as follows (the detailed
derivation steps are deferred to Appendix \ref{sec:Proof-of-Propositions}):
given scalars $0<\alpha<1$ and $0<\beta<1$, the stepsize $\gamma^{t}$
is set to be $\gamma^{t}=\beta^{m_{t}}$, where $m_{t}$ is the smallest
nonnegative integer $m$ satisfying the inequality in (\ref{eq:successive-line-search-proposed})
shown at the top of this page, which is the same as (\ref{eq:successive-line-search-proposed-simplified})
after removing the constants that appear on both sides. Note that
the smooth function $f$ needs to be evaluated several times for $m=1,2,\ldots,m_{t}$
as in traditional successive line search scheme, but we only have
to evaluate the nonsmooth function $g^{+}$ once at $\mathbb{B}\mathbf{x}^{t}$,
i.e., $g^{+}(\mathbb{B}\mathbf{x}^{t})$.\addtocounter{equation}{2}

We show in the following proposition that the stepsize obtained by
the proposed exact/successive line search (\ref{eq:exact-line-search-proposed})
and (\ref{eq:successive-line-search-proposed}) is nonzero, i.e.,
$\gamma^{t}\in(0,1]$ and $h(\mathbf{x}^{t+1})<h(\mathbf{x}^{t})$.
\begin{prop}
[Existence of a nontrivial stepsize]\label{prop:stepsize}If $\mathbb{B}\mathbf{x}^{t}-\mathbf{x}^{t}$
is a descent direction of $\overline{h}(\mathbf{x};\mathbf{x}^{t})$
at the point $\mathbf{x}=\mathbf{x}^{t}$ in the sense of (\ref{eq:descent-direction}),
then the stepsize given by the proposed exact line search (\ref{eq:exact-line-search-proposed})
or the proposed successive line search (\ref{eq:successive-line-search-proposed})
is nonzero, i.e., $\gamma^{t}\in(0,1]$.
\end{prop}
\begin{IEEEproof}
See Appendix \ref{sec:Proof-of-Propositions}.
\end{IEEEproof}
\begin{algorithm}[t]
\textbf{Data: }$t=0$, $\mathbf{x}^{0}$ (arbitrary but fixed, e.g.,
$\mathbf{x}^{0}=\mathbf{0}$), stop criterion $\delta$.

\textbf{S1: }Compute $\mathbb{B}\mathbf{x}^{t}$ according to (\ref{eq:best-response}).

\textbf{S2: }Determine the stepsize $\gamma^{t}$ by the exact line
search (\ref{eq:exact-line-search-proposed}) or the successive line
search (\ref{eq:successive-line-search-proposed}).

\textbf{S3: }Update\textbf{ $\mathbf{x}^{t+1}$} according to (\ref{eq:variable-update}).

\textbf{S4: }If $|(\mathbb{B}\mathbf{x}^{t}-\mathbf{x}^{t})^{T}(\nabla f(\mathbf{x}^{t})-\boldsymbol{\xi}^{-}(\mathbf{x}^{t}))+g^{+}(\mathbb{B}\mathbf{x}^{t})-g^{+}(\mathbf{x}^{t})|\leq\delta$,
STOP; otherwise $t\leftarrow t+1$ and go to \textbf{S1}.

\caption{\label{alg:Successive-approximation-method}The proposed successive
convex approximation framework for problem (\ref{eq:problem-formulation})}
\end{algorithm}

The proposed SCA framework is summarized in Algorithm \ref{alg:Successive-approximation-method}
and its convergence properties are given in the following theorem.
\begin{thm}
[Convergence to a stationary point]\label{thm:convergence}Consider
the sequence $\left\{ \mathbf{x}^{t}\right\} $ generated by Algorithm
\ref{alg:Successive-approximation-method}. Provided that Assumptions
(A1)-(A3) as well as the following assumptions are satisfied:

\begin{enumerate}

\item[\emph{(A4)}] The solution set $\mathcal{S}(\mathbf{x}^{t})$
is nonempty for $t=1,2,\ldots$;

\item[\emph{(A5)}] Given any convergent subsequence $\left\{ \mathbf{x}^{t}\right\} _{t\in\mathcal{T}}$
where $\mathcal{T}\subseteq\left\{ 1,2,\ldots\right\} $, the sequence
$\left\{ \mathbb{B}\mathbf{x}^{t}\right\} _{t\in\mathcal{T}}$ is
bounded.

\end{enumerate}Then any limit point of $\left\{ \mathbf{x}^{t}\right\} $
is a stationary point of (\ref{eq:problem-formulation}).
\end{thm}
\begin{IEEEproof}
See Appendix \ref{sec:Proof-of-Theorem}.
\end{IEEEproof}
Sufficient conditions for Assumptions (A4)-(A5) are that either the
feasible set $\mathcal{X}$ in (\ref{eq:approximate-problem}) is
bounded or the approximate function in (\ref{eq:approximate-problem})
is strongly convex \cite{Robinson1974}. We will show that these assumptions
are satisfied by the example application in the next section.

If, in addition, $\tilde{f}(\mathbf{x};\mathbf{x}^{t})$ in the approximate
function (\ref{eq:approximate-function}) is a global upper bound
of $f(\mathbf{x})$, then the proposed Algorithm \ref{alg:Successive-approximation-method}
converges (in the sense specified by Theorem \ref{thm:convergence})
under a constant unit stepsize $\gamma^{t}=1$. We omit the details
due to the page limit.

In what follows, we draw some comments on the proposed algorithm's
features and connections to existing algorithms.

\textbf{On the choice of approximate function}. Note that different
choices of $\tilde{f}(\mathbf{x};\mathbf{x}^{t})$ lead to different
algorithms. We mention for the self-containedness of this paper two
commonly used approximate functions, and assume for now that the constraint
set $\mathcal{X}$ has a Cartesian product structure and $g^{+}$
is separable, i.e., $g^{+}(\mathbf{x})=\sum_{k=1}^{K}g^{+}(\mathbf{x}_{k})$.
We refer the interested readers to \cite[Sec. III-B]{Yang_ConvexApprox}
for a more comprehensive discussion.

\emph{Proximal type approximation.} The proximal type approximate
function $\tilde{h}(\mathbf{x};\mathbf{x}^{t})$ has the following
form \cite[Sec. 4.2]{Parikh2014}:\begin{subequations}\label{eq:proximal-approximation}
\begin{equation}
\underbrace{f(\mathbf{x}^{t})+\nabla f(\mathbf{x}^{t})(\mathbf{x}-\mathbf{x}^{t})+\frac{c^{t}}{2}\bigl\Vert\mathbf{x}-\mathbf{x}^{t}\bigr\Vert^{2}}_{\tilde{f}(\mathbf{x};\mathbf{x}^{t})}-(\mathbf{x}-\mathbf{x}^{t})^{T}\boldsymbol{\xi}^{-}(\mathbf{x}^{t})+g(\mathbf{x})\label{eq:proximal-approximate-function}
\end{equation}
where $c^{t}>0$. Since the approximate function is separable among
the different block variables and the constraint set has a Cartesian
structure, minimizing the approximate function to obtain $\mathbb{B}\mathbf{x}^{t}$
is equivalent to set $\mathbb{B}\mathbf{x}^{t}=(\mathbb{B}_{k}\mathbf{x}^{t})_{k=1}^{K}$
where $\mathbf{x}=(\mathbf{x}_{k})_{k=1}^{K}$ and
\begin{equation}
\mathbb{B}_{k}\mathbf{x}^{t}\triangleq\underset{\mathbf{x}_{k}\in\mathcal{X}_{k}}{\arg\min}\left\{ \begin{array}{l}
\nabla_{k}f(\mathbf{x}^{t})(\mathbf{x}_{k}-\mathbf{x}_{k}^{t})+\frac{c^{t}}{2}\bigl\Vert\mathbf{x}_{k}-\mathbf{x}_{k}^{t}\bigr\Vert^{2}\smallskip\\
-(\mathbf{x}_{k}-\mathbf{x}_{k}^{t})^{T}\boldsymbol{\xi}_{k}^{-}(\mathbf{x}^{t})+g(\mathbf{x}_{k})
\end{array}\right\} ,\label{eq:proximal-approximate-problem}
\end{equation}
\end{subequations}for all $k=1,\ldots,K$. According to Theorem \ref{thm:convergence}
and the discussion that immediately follows, the proposed algorithm
converges under a constant unit stepsize if $\tilde{f}(\mathbf{x};\mathbf{x}^{t})$
in (\ref{eq:proximal-approximate-function}) is a global upper bound
of $f(\mathbf{x})$, which is indeed the case when $c^{t}\geq L_{\nabla f}$
($L_{\nabla f}$ is the Lipschitz constant of $\nabla f$) in view
of the descent lemma \cite[Prop. A.24]{bertsekas1999nonlinear}.

\emph{Best-response type approximation.} In problem (\ref{eq:problem-formulation}),
if $f(\mathbf{x})$ is convex in each $\mathbf{x}_{k}$ where $k=1,\ldots,K$
(but not necessarily jointly convex in $(\mathbf{x}_{1},\ldots,\mathbf{x}_{K})$),
the best-response type approximate function is defined as\begin{subequations}\label{eq:jacobi-approximation}
\begin{equation}
\tilde{f}(\mathbf{x};\mathbf{x}^{t})={\textstyle \sum_{k=1}^{K}}f(\mathbf{x}_{k},\mathbf{x}_{-k}^{t}),\label{eq:jacobi-approximate-function}
\end{equation}
and the approximate problem is
\begin{equation}
\mathbb{B}_{k}\mathbf{x}^{t}=\underset{\mathbf{x}_{k}\in\mathcal{X}_{k}}{\arg\min}\Bigl\{ f(\mathbf{x}_{k},\mathbf{x}_{-k}^{t})-(\mathbf{x}_{k}-\mathbf{x}_{k}^{t})^{T}\boldsymbol{\xi}_{k}^{-}(\mathbf{x}^{t})+g(\mathbf{x}_{k})\Bigr\},\label{eq:jacobi-approximate-problem}
\end{equation}
\end{subequations}for all $k=1,\ldots,K$. Comparing (\ref{eq:jacobi-approximation})
with (\ref{eq:proximal-approximation}), we see that the function
$f$ is not linearized in (\ref{eq:jacobi-approximate-problem}).
The best-response type algorithm typically converges faster than the
proximal type algorithm because the desirable property such as convexity
is preserved in the best-response type approximation while it is lost
when $f(\mathbf{x})$ is being linearized in the proximal type approximation.

\textbf{On the proposed line search schemes. }Since the objective
function in the proposed exact line search scheme (\ref{eq:exact-line-search-proposed})
is an upper bound of the objective function in (\ref{eq:exact-line-search-traditional-upper-bound})
(see the discussion after (\ref{eq:exact-line-search-proposed})),
the obtained decrease by the proposed line search $\gamma^{t}$ in
(\ref{eq:exact-line-search-proposed}) is generally smaller than that
of $\gamma_{\textrm{ub}}^{t}$ in (\ref{eq:exact-line-search-traditional-upper-bound}),
the line search over the upper bound function $\overline{h}(\mathbf{x};\mathbf{x}^{t})$,
which is furthermore smaller than that of $\gamma_{\textrm{opt}}^{t}$
in (\ref{eq:exact-line-search-traditional-original}), the line search
over the original function $h(\mathbf{x})$:
\begin{align*}
h(\mathbf{x}^{t}+\gamma_{\textrm{opt}}^{t}(\mathbb{B}\mathbf{x}^{t}-\mathbf{x}^{t})) & \leq h(\mathbf{x}^{t}+\gamma_{\textrm{ub}}^{t}(\mathbb{B}\mathbf{x}^{t}-\mathbf{x}^{t}))\\
 & \leq h(\mathbf{x}^{t}+\gamma^{t}(\mathbb{B}\mathbf{x}^{t}-\mathbf{x}^{t}))<h(\mathbf{x}^{t}).
\end{align*}
Nevertheless, the order of complexity is reversed. To see this, assume
$f(\mathbf{x})$ is convex. Then the optimization problem in (\ref{eq:exact-line-search-traditional-original}),
(\ref{eq:exact-line-search-traditional-upper-bound}), and (\ref{eq:exact-line-search-proposed})
is nonconvex and nondifferentiable, convex but nondifferentiable,
and convex and differentiable, respectively. We will illustrate later
by several example applications that the proposed line search scheme
achieves a good tradeoff between complexity and speed.

\textbf{On the convergence speed of the proposed algorithm.} The proposed
algorithm presents a fast convergence behavior because we could choose
the approximate function so that the problem structure is exploited
to a larger extent, for example, the partial convexity in the best-response
type approximation. Furthermore, the line search leads to a much faster
convergence than predetermined stepsizes such as constant stepsizes
and decreasing stepsizes.

\textbf{On the complexity of the proposed algorithm.} The Algorithm
\ref{alg:Successive-approximation-method} has a low complexity due
to the use of an approximate function and the line search scheme over
a differentiable function. The benefits of employing the approximate
function $\tilde{f}(\mathbf{x};\mathbf{x}^{t})$ are twofold. On the
one hand, it is a convex function by Assumption (A1), so the approximate
problem (\ref{eq:approximate-problem}) is a convex problem, which
is presumably easier to solve than (\ref{eq:CCCP}) which is nonconvex
if $f(\mathbf{x})$ is nonconvex. On the other hand, it can be tailored
according to the structure of the problem at hand so that the approximate
problem (\ref{eq:approximate-problem}) is even easier to solve. For
example, if $g^{+}(\mathbf{x})$ is separable among the scalar elements
of $\mathbf{x}$ (as in, e.g., $\ell_{1}$-norm $\left\Vert \mathbf{x}\right\Vert _{1}=\sum_{k=1}^{K}|x_{k}|$),
we can choose $\tilde{f}(\mathbf{x};\mathbf{x}^{t})$ to be separable
as well, so that the problem (\ref{eq:approximate-problem}) can be
decomposed into independent subproblems which are then solved in parallel.
Furthermore, the proposed line search scheme (\ref{eq:exact-line-search-proposed})
is carried out over a differentiable function, which is presumably
much easier to implement than traditional schemes (\ref{eq:exact-line-search-traditional})
over nonconvex nonsmooth functions.

\textbf{On the connection to the classic MM method \cite{Gasso2009}.}
Assume $f$ is convex.\footnote{This is an assumption made in \cite{Gasso2009}.}
The proposed algorithm includes as a special case the MM method proposed
in \cite{Gasso2009} by setting $\tilde{f}(\mathbf{x};\mathbf{x}^{t})=f(\mathbf{x})$,
i.e., no approximation is employed. For this particular choice of
approximate function, it can be verified that the assumptions (A1)-(A3)
are satisfied. Interpreting the MM method as a special case of the
proposed algorithm consolidates once more that choosing an approximate
function that is easier to optimize may notably reduce the algorithm's
complexity.

\textbf{On the connection to the SCA framework for a convex $g$ \cite{Yang_ConvexApprox}.
}The proposed framework includes as a special case the SCA framework
proposed in \cite{Yang_ConvexApprox} for a convex $g$: assume $g^{-}(\mathbf{x})=0$,
and the approximate function in (\ref{eq:approximate-function}) reduces
to $\tilde{h}(\mathbf{x};\mathbf{x}^{t})=\tilde{f}(\mathbf{x};\mathbf{x}^{t})+g^{+}(\mathbf{x})$.

\textbf{On the connection to the GIST algorithm }\cite{Yang2015_arxiv}\textbf{.
}Assume $g^{-}(\mathbf{x})=0$.\footnote{This is an assumption made in \cite{Yang2015_arxiv}.}
In the GIST algorithm \cite{Yang2015_arxiv}, the variable is updated
as follows:
\begin{equation}
\mathbf{x}^{t+1}=\underset{\mathbf{x}}{\arg\min}\;(\mathbf{x}-\mathbf{x}^{t})\nabla f(\mathbf{x}^{t})+\frac{c^{t}}{2}\left\Vert \mathbf{x}-\mathbf{x}^{t}\right\Vert ^{2}+g^{+}(\mathbf{x}).\label{eq:proximal-gradient}
\end{equation}
This is a special case of the proximal type algorithm by choosing
$c^{t}\geq L_{\nabla f}$ and $\gamma^{t}=1$. When the value of $L_{\nabla f}$
is unknown, $c^{t}$ is estimated iteratively: for a constant $\beta\in(0,1)$,
define $\mathbf{x}^{\star}(\beta^{m})$ as
\begin{equation}
\mathbf{x}^{\star}(\beta^{m})\triangleq\underset{\mathbf{x}}{\arg\min}\;(\mathbf{x}-\mathbf{x}^{t})\nabla f(\mathbf{x}^{t})+\frac{1}{2\beta^{m}}\left\Vert \mathbf{x}-\mathbf{x}^{t}\right\Vert ^{2}+g^{+}(\mathbf{x}).\label{eq:proximal-gradient-line-search-1}
\end{equation}
Then $c^{t}=1/\beta^{m_{t}}$ and $\mathbf{x}^{t+1}=\mathbf{x}^{\star}(\beta^{m_{t}})$
while $m_{t}$ is the smallest nonnegative integer such that the following
inequality is satisfied for some $\alpha\in(0,1)$:
\begin{align*}
 & f(\mathbf{x}^{\star}(\eta^{m_{t}}))+g(\mathbf{x}^{\star}(\eta^{m_{t}}))\\
<\: & f(\mathbf{x}^{t})+g(\mathbf{x}^{t})-\frac{\alpha}{\beta^{m_{t}}}\left\Vert \mathbf{x}^{\star}(\eta^{m_{t}})-\mathbf{x}^{t}\right\Vert ^{2}.
\end{align*}
This implies that, in the GIST algorithm, $\mathbf{x}^{\star}(\beta^{m})$
and $g(\mathbf{x}^{\star}(\beta^{m}))$ are evaluated for $m_{t}+1$
times, namely, $m=0,1,\ldots,m_{t}$. This is however not necessary
in the proposed successive line search (\ref{eq:successive-line-search-proposed}),
because $\mathbb{B}\mathbf{x}^{t}$ given by (\ref{eq:best-response})
does not depend on any unknown parameters and both $\mathbb{B}\mathbf{x}^{t}$
and $g^{+}(\mathbb{B}\mathbf{x}^{t})$ only need to be computed once.
Therefore, the algorithmic complexity could be notably reduced by
employing a convex approximate function that is not necessarily an
upper bound of the original function $h(\mathbf{x})$.

\section{\label{sec:application-rank-minimization}Network Anomaly Detection
Through Sparsity Regularized Rank Minimization}

In this section, we propose an iterative algorithm by customizing
Algorithm \ref{alg:Successive-approximation-method} to solve the
network anomaly detection problem introduced in Sec. \ref{sec:Problem-Formulation}-A.
For the simplicity of cross reference, we duplicate the problem formulation
here
\begin{equation}
\underset{\mathbf{P},\mathbf{Q},\mathbf{S}}{\textrm{minimize}}\;\frac{1}{2}\left\Vert \mathbf{P}\mathbf{Q}+\mathbf{D}\mathbf{S}-\mathbf{Y}\right\Vert _{F}^{2}+\frac{\lambda}{2}\left(\left\Vert \mathbf{P}\right\Vert _{F}^{2}+\left\Vert \mathbf{Q}\right\Vert _{F}^{2}\right)+\mu\left\Vert \mathbf{S}\right\Vert _{1},\label{eq:rank-problem-formulation}
\end{equation}
and remark again that problem (\ref{eq:rank-problem-formulation})
is a special case of (\ref{eq:problem-formulation}) by setting
\begin{align*}
f(\mathbf{P},\mathbf{Q},\mathbf{S}) & \triangleq\frac{1}{2}\left\Vert \mathbf{P}\mathbf{Q}+\mathbf{D}\mathbf{S}-\mathbf{Y}\right\Vert _{F}^{2}+\frac{\lambda}{2}\left(\left\Vert \mathbf{P}\right\Vert _{F}^{2}+\left\Vert \mathbf{Q}\right\Vert _{F}^{2}\right),\\
g(\mathbf{S}) & \triangleq\mu\left\Vert \mathbf{S}\right\Vert _{1},
\end{align*}
where $g(\mathbf{S})$ is convex. To simplify the notation, we use
$\mathbf{Z}$ as a compact notation for $(\mathbf{P},\mathbf{Q},\mathbf{S})$:
$\mathbf{Z}\triangleq(\mathbf{P},\mathbf{Q},\mathbf{S})$; in the
rest of this section, $\mathbf{Z}$ and $(\mathbf{P},\mathbf{Q},\mathbf{S})$
are used interchangeably.

\emph{Related work.} We first briefly describe the BCD algorithm adopted
in \cite{Mardani2013b} to find a stationary point of the nonconvex
problem (\ref{eq:rank-problem-formulation}), where the variables
are updated sequentially according to their best-response. For example,
when $\mathbf{P}$ (or $\mathbf{Q}$) is updated, the variables $(\mathbf{Q,S})$
(or $(\mathbf{P,S})$) are fixed. When $(\mathbf{P,Q})$ is fixed
for example, the optimization problem w.r.t. $\mathbf{S}$ decouples
among its columns:
\begin{align*}
 & \frac{1}{2}\left\Vert \mathbf{P}\mathbf{Q}+\mathbf{D}\mathbf{S}-\mathbf{Y}\right\Vert _{F}^{2}+\mu\left\Vert \mathbf{S}\right\Vert _{1}\\
=\; & \sum_{k=1}^{K}\left(\frac{1}{2}\left\Vert \mathbf{P}\mathbf{q}_{k}-\mathbf{D}\mathbf{s}_{k}-\mathbf{y}_{k}\right\Vert _{2}^{2}+\mu\left\Vert \mathbf{s}_{k}\right\Vert _{1}\right),
\end{align*}
where $\mathbf{q}_{k}$, $\mathbf{s}_{k}$ and $\mathbf{y}_{k}$ is
the $k$-th column of $\mathbf{Q}$, $\mathbf{S}$ and $\mathbf{Y}$,
respectively. However, the optimization problem w.r.t. $\mathbf{s}_{k}$
does not have a closed-form solution and is not easy to solve. To
reduce the complexity, the elements of $\mathbf{S}$ are updated row-wise,
as the optimization problem w.r.t. $s_{i,k}$, the $(i,k)$-th element
of $\mathbf{S}$, has a closed-form solution:
\[
\underset{(s_{i,k})_{k=1}^{K}}{\textrm{minimize}}\;\sum_{k=1}^{K}\left(\begin{array}{l}
\frac{1}{2}\left\Vert \mathbf{P}\mathbf{q}_{k}-\mathbf{d}_{i}s_{i,k}-\sum_{j=1,j\neq i}^{I}\mathbf{d}_{j}s_{j,k}-\mathbf{y}_{k}\right\Vert _{2}^{2}\smallskip\\
+\mu|s_{i,k}|+\mu\sum_{j=1,j\neq i}^{K}|s_{j,k}|
\end{array}\right),
\]
where $\mathbf{d}_{i}$ is the $i$-th column of $\mathbf{D}$, and
$s_{i,k}$ is the $i$-th element of $\mathbf{s}_{k}$ (and hence
the $(j,k)$-th element of $\mathbf{S}$). Solving the above optimization
problem w.r.t. $(s_{i,k})_{k=1}^{K}$ for a given $i$ results in
simultaneous update of all elements in the same ($i$-th) row of $\mathbf{S}$,
and changing $i$ from iteration to iteration results in the sequential
row-wise update. Nevertheless, a major drawback of the sequential
row-wise update is that it may incur a large delay because the $(i+1)$-th
row cannot be updated until the $i$-th row is updated and the delay
may be very large when $I$, the number of rows, is large, which is
a norm rather than an exception in big data analytics \cite{Slavakis2014a}.

\emph{Proposed algorithm.}\textbf{ }Although $f(\mathbf{P,Q,S})$
in (\ref{eq:rank-problem-formulation}) is not jointly convex w.r.t.
$(\mathbf{P,Q,S})$, it is individual convex in $\mathbf{P}$, $\mathbf{Q}$
and $\mathbf{S}$. In other words, $f(\mathbf{P,Q,S})$ is convex
w.r.t. one variable while the other two variables are fixed. This
leads to the best-response type approximation: given $\mathbf{Z}^{t}=(\mathbf{P}^{t},\mathbf{Q}^{t},\mathbf{S}^{t})$
in iteration $t$, we approximate the original nonconvex function
$f(\mathbf{Z})$ by a convex function $\tilde{f}(\mathbf{Z};\mathbf{Z}^{t})$
that is of the following form,
\begin{equation}
\tilde{f}(\mathbf{Z};\mathbf{Z}^{t})=\tilde{f}_{P}(\mathbf{P};\mathbf{Z}^{t})+\tilde{f}_{Q}(\mathbf{Q};\mathbf{Z}^{t})+\tilde{f}_{S}(\mathbf{S};\mathbf{Z}^{t}),\label{eq:rank-approximate-function}
\end{equation}
where\begin{subequations}\label{eq:rank-approximate-function-individual}
\begin{equation}
\tilde{f}_{P}(\mathbf{P};\mathbf{Z}^{t})\triangleq f(\mathbf{P},\mathbf{Q}^{t},\mathbf{S}^{t})=\frac{1}{2}\left\Vert \mathbf{P}\mathbf{Q}^{t}+\mathbf{D}\mathbf{S}^{t}-\mathbf{Y}\right\Vert _{F}^{2}+\frac{\lambda}{2}\left\Vert \mathbf{P}\right\Vert _{F}^{2},\label{eq:rank-approximate-function-P}
\end{equation}
\begin{equation}
\tilde{f}_{Q}(\mathbf{Q};\mathbf{Z}^{t})\triangleq f(\mathbf{P}^{t},\mathbf{Q},\mathbf{S}^{t})=\frac{1}{2}\left\Vert \mathbf{P}^{t}\mathbf{Q}+\mathbf{D}\mathbf{S}^{t}-\mathbf{Y}\right\Vert _{F}^{2}+\frac{\lambda}{2}\left\Vert \mathbf{Q}\right\Vert _{F}^{2},\label{eq:rank-approximate-function-Q}
\end{equation}
\begin{align}
\tilde{f}_{S}(\mathbf{S};\mathbf{Z}^{t}) & \triangleq\sum_{i,k}f(\mathbf{P}^{t},\mathbf{Q}^{t},s_{i,k},(s_{j,k}^{t})_{j\neq i},(\mathbf{s}_{j}^{t})_{j\neq i})\nonumber \\
 & =\sum_{i,k}\frac{1}{2}\left\Vert \mathbf{P}^{t}\mathbf{q}_{k}^{t}+\mathbf{d}_{i}s_{i,k}+{\textstyle \sum_{j\neq i}}\mathbf{d}_{j}s_{j,k}^{t}-\mathbf{y}_{k}\right\Vert _{2}^{2}\nonumber \\
 & =\textrm{tr}(\mathbf{S}^{T}\mathbf{d}(\mathbf{D}^{T}\mathbf{D})\mathbf{S})\nonumber \\
 & \qquad-\textrm{tr}(\mathbf{S}^{T}(\mathbf{d}(\mathbf{D}^{T}\mathbf{D})\mathbf{S}^{t}-\mathbf{D}^{T}(\mathbf{D}\mathbf{S}^{t}-\mathbf{Y}+\mathbf{P}^{t}\mathbf{Q}^{t}))),\label{eq:rank-approximate-function-S}
\end{align}
\end{subequations}with $\mathbf{q}_{k}$ (or $\mathbf{y}_{k}$) and
$\mathbf{d}_{i}$ denoting the $k$-th and $i$-th column of $\mathbf{Q}$
(or $\mathbf{Y}$) and $\mathbf{D}$, respectively, while $\mathbf{d}(\mathbf{D}^{T}\mathbf{D})$
denotes a diagonal matrix with elements on the main diagonal identical
to those of the matrix $\mathbf{D}^{T}\mathbf{D}$. Note that in the
approximate function w.r.t. $\mathbf{P}$ and $\mathbf{Q}$, the remaining
variables $(\mathbf{Q,S})$ and $(\mathbf{P,S})$ are fixed, respectively.
Although it is tempting to define the approximate function of $f(\mathbf{P,Q,S})$
w.r.t. $\mathbf{S}$ by fixing $\mathbf{P}$ and $\mathbf{Q}$, minimizing
$f(\mathbf{P}^{t},\mathbf{Q}^{t},\mathbf{S})$ w.r.t. the matrix variable
$\mathbf{S}$ does not have a closed-form solution and must be solved
iteratively. Therefore the proposed approximate function $\tilde{f}_{S}(\mathbf{S};\mathbf{Z}^{t})$
in (\ref{eq:rank-approximate-function-S}) consists of $IK$ component
functions, and in the $(i,k)$-th component function, $s_{i,k}$ is
the variable while all other variables are fixed, namely, $\mathbf{P}$,
$\mathbf{Q}$, $(s_{j,k})_{j\neq i}$, and $(\mathbf{s}_{j})_{j\neq i}$.
As we will show shortly, minimizing $\tilde{f}(\mathbf{S};\mathbf{Z}^{t})$
w.r.t. $\mathbf{S}$ exhibits a closed-form solution.

We remark that the approximate function $\tilde{f}(\mathbf{Z;Z}^{t})$
is a (strongly) convex function and it is differentiable in both $\mathbf{Z}$
and $\mathbf{Z}^{t}$. Furthermore, the gradient of the approximate
function $\tilde{f}(\mathbf{P,Q,S};\mathbf{Z}^{t})$ is equal to that
of $f(\mathbf{P,Q,S})$ at $\mathbf{Z}=\mathbf{Z}^{t}$. To see this:
\[
\nabla_{\mathbf{P}}\tilde{f}(\mathbf{Z};\mathbf{Z}^{t})=\nabla_{\mathbf{P}}\tilde{f}_{P}(\mathbf{P};\mathbf{Z}^{t})=\nabla_{\mathbf{P}}\left.f(\mathbf{P},\mathbf{Q}^{t},\mathbf{S}^{t})\right|_{\mathbf{P=P}^{t}},
\]
and similarly $\nabla_{\mathbf{Q}}\tilde{f}(\mathbf{Z};\mathbf{Z}^{t})=\left.\nabla_{\mathbf{Q}}f(\mathbf{P,Q,S})\right|_{\mathbf{Z}=\mathbf{Z}^{t}}$.
Furthermore, $\nabla_{\mathbf{S}}\tilde{f}(\mathbf{Z};\mathbf{Z}^{t})=(\nabla_{s_{i,k}}\tilde{f}(\mathbf{Z};\mathbf{Z}^{t}))_{i,k}$
while
\begin{align*}
\nabla_{s_{i,k}}\tilde{f}(\mathbf{Z};\mathbf{Z}^{t}) & =\nabla_{s_{i,k}}\tilde{f}_{S}(\mathbf{S};\mathbf{Z}^{t})\\
 & =\nabla_{s_{i,k}}f(\mathbf{P}^{t},\mathbf{Q}^{t},s_{i,k},\mathbf{s}_{i,-k}^{t},\mathbf{s}_{-i}^{t})\\
 & =\left.\nabla_{s_{i,k}}f(\mathbf{P,Q,S})\right|_{\mathbf{Z}=\mathbf{Z}^{t}}.
\end{align*}
Therefore Assumptions (A1)-(A3) are satisfied.

In iteration $t$, the approximate problem consists of minimizing
the approximate function:
\begin{align}
\underset{\mathbf{Z}=(\mathbf{P},\mathbf{Q},\mathbf{S})}{\textrm{minimize}} & \quad\underbrace{\tilde{f}_{P}(\mathbf{P};\mathbf{Z}^{t})+\tilde{f}_{Q}(\mathbf{Q};\mathbf{Z}^{t})+\tilde{f}_{S}(\mathbf{S};\mathbf{Z}^{t})}_{\tilde{f}(\mathbf{Z};\mathbf{Z}^{t})}+g(\mathbf{S}).\label{eq:rank-approximate-problem}
\end{align}
Since $\tilde{f}(\mathbf{Z};\mathbf{Z}^{t})$ is strongly convex in
$\mathbf{Z}$ and $g(\mathbf{S})$ is a convex function w.r.t. $\mathbf{S}$,
the approximate problem (\ref{eq:rank-approximate-problem}) is strongly
convex and it has a unique globally optimal solution, which is denoted
as $\mathbb{B}\mathbf{Z}^{t}=(\mathbb{B}_{P}\mathbf{Z}^{t},\mathbb{B}_{Q}\mathbf{Z}^{t},\mathbb{B}_{S}\mathbf{Z}^{t})$.
As the approximate problem (\ref{eq:rank-approximate-problem}) is
separable among the optimization variables $\mathbf{P}$, $\mathbf{Q}$
and $\mathbf{S}$, it naturally decomposes into several smaller problems
which can be solved in parallel:\begin{subequations}\label{eq:rank-approximate-problem-solution}
\begin{align}
\mathbb{B}_{P}\mathbf{Z}^{t} & \triangleq\underset{\mathbf{P}_{k}}{\arg\min}\;\tilde{f}_{P}(\mathbf{P};\mathbf{Z}^{t})\nonumber \\
 & =(\mathbf{Y}-\mathbf{D}\mathbf{S}^{t})(\mathbf{Q}^{t})^{T}(\mathbf{Q}^{t}(\mathbf{Q}^{t})^{T}+\lambda\mathbf{I})^{-1},\label{eq:rank-BP}
\end{align}
\begin{align}
\mathbb{B}_{Q}\mathbf{Z}^{t} & \triangleq\underset{\mathbf{Q}}{\arg\min}\;\tilde{f}_{Q}(\mathbf{Q};\mathbf{Z}^{t})\nonumber \\
 & =((\mathbf{P}^{t})^{T}\mathbf{P}^{t}+\lambda\mathbf{I})^{-1}(\mathbf{P}^{t})^{T}(\mathbf{Y}-\mathbf{D}\mathbf{S}^{t}),\label{eq:rank-BQ}
\end{align}
\begin{align}
\mathbb{B}_{S}\mathbf{Z}^{t} & \triangleq\underset{\mathbf{S}}{\arg\min}\;\tilde{f}_{S}(\mathbf{S};\mathbf{Z}^{t})+g(\mathbf{S})\nonumber \\
 & =\mathbf{d}(\mathbf{D}^{T}\mathbf{D})^{-1}\cdot\nonumber \\
 & \qquad\mathcal{S}_{\mu}\left(\mathbf{d}(\mathbf{D}^{T}\mathbf{D})\mathbf{S}^{t}-\mathbf{D}^{T}(\mathbf{D}\mathbf{S}^{t}-\mathbf{Y}^{t}+\mathbf{P}^{t}\mathbf{Q}^{t})\right),\label{eq:rank-BS}
\end{align}
\end{subequations}where $\mathcal{S}_{\mu}(\mathbf{X})$ is an element-wise
soft-thresholding operator: the $(i,j)$-th element of $\mathcal{S}_{\mu}(\mathbf{X})$
is $[X_{ij}-\lambda]^{+}-[-X_{ij}-\lambda]^{+}$. As we can readily
see from (\ref{eq:rank-approximate-problem-solution}), the approximate
problems can be solved efficiently because the optimal solutions are
provided in an analytical expression.

Since $\tilde{f}(\mathbf{Z};\mathbf{Z}^{t})$ is convex in $\mathbf{Z}$
and differentiable in both $\mathbf{Z}$ and $\mathbf{Z}^{t}$, and
has the same gradient as $f(\mathbf{Z})$ at $\mathbf{Z}=\mathbf{Z}^{t}$,
it follows from Proposition \ref{prop:descent-property} that $\mathbb{B}\mathbf{Z}^{t}-\mathbf{Z}^{t}$
is a descent direction of the original objective function $f(\mathbf{Z})+g(\mathbf{S})$
at $\mathbf{Z}=\mathbf{Z}^{t}$. The variable update in the $t$-th
iteration is thus defined as follows:\begin{subequations}\label{eq:rank-variable-update}
\begin{align}
\mathbf{P}^{t+1} & =\mathbf{P}^{t}+\gamma(\mathbb{B}_{P}\mathbf{Z}^{t}-\mathbf{P}^{t}),\label{eq:rank-variable-update-P}\\
\mathbf{Q}^{t+1} & =\mathbf{Q}^{t}+\gamma(\mathbb{B}_{Q}\mathbf{Z}^{t}-\mathbf{Q}^{t}),\label{eq:rank-variable-update-Q}\\
\mathbf{S}^{t+1} & =\mathbf{S}^{t}+\gamma(\mathbb{B}_{S}\mathbf{Z}^{t}-\mathbf{S}^{t}),\label{eq:rank-variable-update-S}
\end{align}
\end{subequations}where $\gamma\in(0,1]$ is the stepsize that should
be properly selected.

We determine the stepsize $\gamma$ by the proposed exact line search
scheme (\ref{eq:exact-line-search-proposed}):
\begin{equation}
f(\mathbf{Z}^{t}+\gamma(\mathbb{B}\mathbf{Z}^{t}-\mathbf{Z}^{t}))+g(\mathbf{S}^{t})+\gamma(g(\mathbb{B}_{S}\mathbf{Z}^{t})-g(\mathbf{S}^{t})).\label{eq:rank-line-search-proposed-concept}
\end{equation}
After substituting the expressions of $f(\mathbf{Z})$ and $g(\mathbf{S})$
into (\ref{eq:rank-line-search-proposed-concept}), the exact line
search consists in minimizing a fourth order polynomial over the interval
$[0,1]$:
\begin{align}
\gamma^{t} & =\underset{0\leq\gamma\leq1}{\arg\min}\left\{ f(\mathbf{Z}^{t}+\gamma(\mathbb{B}\mathbf{Z}^{t}-\mathbf{Z}^{t}))+\gamma(g(\mathbb{B}_{S}\mathbf{X}^{t})-g(\mathbf{S}^{t}))\right\} \nonumber \\
 & =\underset{0\leq\gamma\leq1}{\arg\min}\left\{ \frac{1}{4}a\gamma^{4}+\frac{1}{3}b\gamma^{3}+\frac{1}{2}c\gamma^{2}+d\gamma\right\} ,\label{eq:rank-line-search-proposed}
\end{align}
where
\begin{align*}
a & \triangleq2\left\Vert \triangle\mathbf{P}^{t}\triangle\mathbf{Q}^{t}\right\Vert _{F}^{2},\\
b & \triangleq3\textrm{tr}(\triangle\mathbf{P}^{t}\triangle\mathbf{Q}^{t}(\mathbf{P}^{t}\triangle\mathbf{Q}^{t}+\triangle\mathbf{P}^{t}\mathbf{Q}^{t}+\mathbf{D}\triangle\mathbf{S}^{t})^{T}),\\
c & \triangleq2\textrm{tr}(\triangle\mathbf{P}^{t}\triangle\mathbf{Q}^{t}(\mathbf{P}^{t}\mathbf{Q}^{t}+\mathbf{D}\mathbf{S}^{t}-\mathbf{Y}^{t})^{T})\\
 & \quad+\left\Vert \mathbf{P}^{t}\triangle\mathbf{Q}^{t}+\triangle\mathbf{P}^{t}\mathbf{Q}^{t}+\mathbf{D}\triangle\mathbf{S}^{t}\right\Vert _{F}^{2}\\
 & \quad+\lambda(\left\Vert \triangle\mathbf{P}^{t}\right\Vert _{F}^{2}+\left\Vert \triangle\mathbf{Q}^{t}\right\Vert _{F}^{2}),\\
d & \triangleq\textrm{tr}((\mathbf{P}^{t}\triangle\mathbf{Q}^{t}+\triangle\mathbf{P}^{t}\mathbf{Q}^{t}+\mathbf{D}\triangle\mathbf{S}^{t})(\mathbf{P}^{t}\mathbf{Q}^{t}+\mathbf{D}\mathbf{S}^{t}-\mathbf{Y}^{t}))\\
 & \quad+\lambda(\textrm{tr}(\mathbf{P}^{t}\triangle\mathbf{P}^{t})+\textrm{tr}(\mathbf{Q}^{t}\triangle\mathbf{Q}^{t}))+\mu(\left\Vert \mathbb{B}_{S}\mathbf{X}^{t}\right\Vert _{1}-\left\Vert \mathbf{S}^{t}\right\Vert _{1}),
\end{align*}
for $\triangle\mathbf{P}^{t}\triangleq\mathbb{B}_{P}\mathbf{Z}^{t}-\mathbf{P}^{t}$,
$\triangle\mathbf{Q}^{t}\triangleq\mathbb{B}_{Q}\mathbf{Z}^{t}-\mathbf{Q}^{t}$
and $\triangle\mathbf{S}^{t}\triangleq\mathbb{B}_{S}\mathbf{Z}^{t}-\mathbf{S}^{t}$.
Finding the optimal points of (\ref{eq:rank-line-search-proposed})
is equivalent to finding the nonnegative real root of a third-order
polynomial. Making use of Cardano's method, we write $\gamma^{t}$
defined in (\ref{eq:rank-line-search-proposed}) as the closed-form
expression:\begin{subequations}\label{eq:rank-line-search-proposed-closed-form}
\begin{align}
\gamma^{t} & =[\bar{\gamma}^{t}]_{0}^{1},\label{eq:rank-line-search-proposed-closed-form-1}\\
\bar{\gamma}^{t} & =\sqrt[3]{\Sigma_{1}+\sqrt{\Sigma_{1}^{2}+\Sigma_{2}^{3}}}+\sqrt[3]{\Sigma_{1}-\sqrt{\Sigma_{1}^{2}+\Sigma_{2}^{3}}}-\frac{b}{3a},\label{eq:rank-line-search-proposed-closed-form-2}
\end{align}
\end{subequations}where $\left[x\right]_{0}^{1}=\max(\min(x,1),0)$
is the projection of $x$ onto the interval $[0,1]$, $\Sigma_{1}\triangleq-(b/3a)^{3}+bc/6a^{2}-d/2a$
and $\Sigma_{2}\triangleq c/3a-(b/3a)^{2}$. Note that in (\ref{eq:rank-line-search-proposed-closed-form-2}),
the right hand side contains three values (two of them can attain
complex numbers), and the equal sign must be interpreted as assigning
the smallest real nonnegative values.

\begin{algorithm}[t]
\textbf{Data: }$t=0$, $\mathbf{Z}^{0}$ (arbitrary but fixed), stop
criterion $\delta$.

\textbf{S1: }Compute $(\mathbb{B}_{P}\mathbf{Z}^{t},\mathbb{B}_{Q}\mathbf{Z}^{t},\mathbb{B}_{S}\mathbf{Z}^{t})$
according to (\ref{eq:rank-approximate-problem-solution}).

\textbf{S2: }Determine the stepsize $\gamma^{t}$ by the exact line
search (\ref{eq:rank-line-search-proposed-closed-form}).

\textbf{S3: }Update $(\mathbf{P},\mathbf{Q},\mathbf{Z})$ according
to (\ref{eq:rank-variable-update}).

\textbf{S4: }If $\left|\textrm{tr}((\mathbb{B}\mathbf{Z}^{t}-\mathbf{Z}^{t})^{T}\nabla f(\mathbf{Z}^{t}))+g(\mathbb{B}_{S}\mathbf{Z}^{t})-g(\mathbf{S}^{t})\right|\leq\delta$,
STOP; otherwise $t\leftarrow t+1$ and go to \textbf{S1}.

\caption{\label{alg:Successive-approximation-method-rank-minimization}STELA:
The proposed parallel best-response with exact line search algorithm
for the sparsity regularized rank minimization problem (\ref{eq:rank-problem-formulation})}
\end{algorithm}

The proposed algorithm is summarized in Algorithm \ref{alg:Successive-approximation-method-rank-minimization},
which we name as the Soft-Thresholding with Exact Line search Algorithm
(STELA). We draw a few comments on its attractive features and compare
it with state-of-the-art algorithms proposed for problem (\ref{eq:rank-problem-formulation}).

\emph{i) Fast convergence.} In each iteration, the variables $\mathbf{P}$,
$\mathbf{Q}$, and $\mathbf{S}$ are updated simultaneously based
on the best-response. The improvement in convergence speed w.r.t.
the BCD algorithm in \cite{Mardani2013b} is notable because in the
BCD algorithm, the optimization w.r.t. each row of $\mathbf{S}$ is
implemented in a sequential order, and the number of rows is usually
very large in big data applications. To avoid the meticulous choice
of stepsizes and further accelerate the convergence, the stepsize
is calculated by the exact line search and it yields faster convergence
than SCA algorithms with diminishing stepsizes \cite{Razaviyayn2014,Scutari_BigData}.

\emph{ii) Low complexity.} The proposed algorithm STELA has a very
low complexity, because both the best-responses $(\mathbb{B}_{P}\mathbf{Z}^{t},\mathbb{B}_{Q}\mathbf{Z}^{t},\mathbb{B}_{S}\mathbf{Z}^{t})$
and the exact line search can be computed by closed-form expressions,
cf. (\ref{eq:rank-approximate-function-individual}) and (\ref{eq:rank-line-search-proposed-closed-form}).
Note that computing $\mathbb{B}_{P}\mathbf{Z}^{t}$ and $\mathbb{B}_{Q}\mathbf{Z}^{t}$
according to (\ref{eq:rank-BP})-(\ref{eq:rank-BQ}) involves a matrix
inverse. This is usually affordable because the matrices to be inverted
are of a dimension $\rho\times\rho$ while $\rho$ is usually small.
Furthermore, the matrix inverse operation could be saved by adopting
an element-wise decomposition for $\mathbf{P}$ and $\mathbf{Q}$
that is in the same essence as $\mathbf{S}$ in (\ref{eq:rank-approximate-function-S}).

\emph{iii) Guaranteed convergence.} In contrast to the ADMM algorithm
\cite{Mardani2013}, the proposed algorithm STELA has a guaranteed
convergence in the sense that every limit point of the sequence $\{\mathbf{Z}^{t}\}_{t}$
is a stationary point of problem (\ref{eq:rank-problem-formulation}).

\subsection{Parallel Decomposition and Implementation of the Proposed Algorithm
STELA}

The proposed algorithm STELA can be further decomposed to enable the
parallel processing over a number of $L$ nodes in a distributed network.
To see this, we first decompose the system model across the nodes:
\[
\mathbf{Y}_{l}=\mathbf{X}_{l}+\mathbf{D}_{l}\mathbf{S}+\mathbf{V}_{l},l=1,\ldots,L,
\]
where $\mathbf{Y}_{l}\in\mathbb{R}^{N_{l}\times K}$, $\mathbf{X}_{l}\in\mathbb{R}^{N_{l}\times K}$,
$\mathbf{D}_{l}\in\mathbb{R}^{N_{l}\times I}$ and $\mathbf{V}_{l}\in\mathbb{R}^{N_{l}\times K}$
consists of $N_{l}$ rows of $\mathbf{Y}$, $\mathbf{X}$, $\mathbf{D}$
and $\mathbf{V}$, respectively:
\[
\mathbf{Y}=\left[\begin{array}{c}
\mathbf{Y}_{1}\\
\mathbf{Y}_{2}\\
\vdots\\
\mathbf{Y}_{L}
\end{array}\right],\mathbf{X}=\left[\begin{array}{c}
\mathbf{X}_{1}\\
\mathbf{X}_{2}\\
\vdots\\
\mathbf{X}_{L}
\end{array}\right],\mathbf{D}=\left[\begin{array}{c}
\mathbf{D}_{1}\\
\mathbf{D}_{2}\\
\vdots\\
\mathbf{D}_{L}
\end{array}\right],\mathbf{V}=\left[\begin{array}{c}
\mathbf{V}_{1}\\
\mathbf{V}_{2}\\
\vdots\\
\mathbf{V}_{L}
\end{array}\right].
\]
Since the variables of interest for the node $l$ are $\mathbf{X}_{k}$
and $\mathbf{S}$, we decompose $\mathbf{P}$ into multiple blocks
$(\mathbf{P}_{l})_{l=1}^{L}$ with $\mathbf{P}_{l}\in\mathbb{R}^{N_{l}\times\rho}$:
\[
\mathbf{P}=\left[\begin{array}{c}
\mathbf{P}_{1}\\
\mathbf{P}_{2}\\
\vdots\\
\mathbf{P}_{L}
\end{array}\right].
\]
All nodes should have access to the variable $\mathbf{Q}$ so that
$\mathbf{X}_{l}$ can be estimated locally by $\mathbf{X}_{l}=\mathbf{P}_{l}\mathbf{Q}$.

The computation of $\mathbb{B}_{P}\mathbf{Z}^{t}$ in (\ref{eq:rank-variable-update-P})
can be decomposed as $\mathbb{B}_{P}\mathbf{Z}^{t}=(\mathbb{B}_{P,l}\mathbf{Z}^{t})_{l=1}^{L}$:
\[
\mathbb{B}_{P,l}\mathbf{Z}^{t}=(\mathbf{Y}_{l}-\mathbf{D}_{l}\mathbf{S}^{t})(\mathbf{Q}^{t})^{T}(\mathbf{Q}^{t}(\mathbf{Q}^{t})^{T}+\lambda\mathbf{I})^{-1},l=1,\ldots,L.
\]
Accordingly, the computation of $\mathbb{B}_{Q}\mathbf{Z}^{t}$ and
$\mathbb{B}_{S}\mathbf{Z}^{t}$ in (\ref{eq:rank-variable-update-Q})
and (\ref{eq:rank-variable-update-S}) can be rewritten as
\begin{align*}
\mathbb{B}_{Q}\mathbf{Z}^{t} & =\left({\textstyle \sum_{l=1}^{L}}(\mathbf{P}_{l}^{t})^{T}\mathbf{P}_{l}^{t}+\lambda\mathbf{I}\right)^{-1}\left({\textstyle \sum_{l=1}^{L}}(\mathbf{P}_{l}^{t})^{T}(\mathbf{Y}_{l}-\mathbf{D}_{l}\mathbf{S}^{t})\right),
\end{align*}
\begin{align*}
\mathbb{B}_{S}\mathbf{Z}^{t} & =\mathbf{d}\left({\textstyle \sum_{l=1}^{L}}\mathbf{D}_{l}^{T}\mathbf{D}_{l}\right)^{-1}\cdot\\
\mathcal{S}_{\mu} & \left(\mathbf{d}\left({\textstyle \sum_{l=1}^{L}}\mathbf{D}_{l}^{T}\mathbf{D}_{l}\right)\mathbf{S}^{t}-{\textstyle \sum_{l=1}^{L}}\mathbf{D}_{l}^{T}(\mathbf{D}_{l}\mathbf{S}^{t}-\mathbf{Y}_{l}^{t}+\mathbf{P}_{l}^{t}\mathbf{Q}^{t})\right).
\end{align*}
Before determining the stepsize, the computation of $a$ in (\ref{eq:rank-line-search-proposed-closed-form})
can also be decomposed among the nodes as $a=\sum_{l=1}^{L}a_{l}$,
where
\[
a_{l}\triangleq2\left\Vert \triangle\mathbf{P}_{l}^{t}\triangle\mathbf{Q}^{t}\right\Vert _{F}^{2}.
\]
The decomposition of $b$, $c$, and $d$ is similar to that of $a$,
where
\begin{align*}
b_{l} & \triangleq3\textrm{tr}(\triangle\mathbf{P}_{l}^{t}\triangle\mathbf{Q}^{t}(\mathbf{P}_{l}^{t}\triangle\mathbf{Q}^{t}+\triangle\mathbf{P}_{l}^{t}\mathbf{Q}^{t}+\mathbf{D}_{l}\triangle\mathbf{S}^{t})^{T}),\\
c_{l} & \triangleq2\textrm{tr}(\triangle\mathbf{P}_{l}^{t}\triangle\mathbf{Q}^{t}(\mathbf{P}_{l}^{t}\mathbf{Q}^{t}+\mathbf{D}_{l}\mathbf{S}^{t}-\mathbf{Y}_{l}^{t})^{T})\\
 & \quad+\left\Vert \mathbf{P}_{l}^{t}\triangle\mathbf{Q}^{t}+\triangle\mathbf{P}_{l}^{t}\mathbf{Q}^{t}+\mathbf{D}_{l}\triangle\mathbf{S}^{t}\right\Vert _{F}^{2}\\
 & \quad+\lambda\left\Vert \triangle\mathbf{P}_{l}^{t}\right\Vert _{F}^{2}+\frac{\lambda}{I}\left\Vert \triangle\mathbf{Q}_{l}^{t}\right\Vert _{F}^{2},\\
d_{l} & \triangleq\textrm{tr}((\mathbf{P}_{l}^{t}\triangle\mathbf{Q}^{t}+\triangle\mathbf{P}_{l}^{t}\mathbf{Q}^{t}+\mathbf{D}_{l}\triangle\mathbf{S}^{t})(\mathbf{P}_{l}^{t}\mathbf{Q}^{t}+\mathbf{D}_{l}\mathbf{S}^{t}-\mathbf{Y}_{l}^{t}))\\
 & \quad+\lambda\textrm{tr}(\mathbf{P}_{l}^{t}\triangle\mathbf{P}_{l}^{t})+\frac{\lambda}{I}\textrm{tr}(\mathbf{Q}^{t}\triangle\mathbf{Q}^{t})+\frac{\mu}{I}(\left\Vert \mathbb{B}_{S}\mathbf{X}^{t}\right\Vert _{1}-\left\Vert \mathbf{S}^{t}\right\Vert _{1}).
\end{align*}
To compute the stepsize as in (\ref{eq:rank-line-search-proposed-closed-form}),
the nodes mutually exchange $(a_{l},b_{l},c_{l},d_{l})$. The four
dimensional vector $(a_{l},b_{l},c_{l},d_{l})$ provides each node
with all the necessary information to individually calculate $(a,b,c,d)$
and $(\Sigma_{1},\Sigma_{2},\Sigma_{3})$, and then the stepsize $\gamma^{t}$
according to (\ref{eq:rank-line-search-proposed-closed-form}). The
signaling incurred by the exact line search is thus small and affordable.

\subsection{Numerical Simulations}

\begin{figure}[t]
\center \includegraphics[scale=0.65]{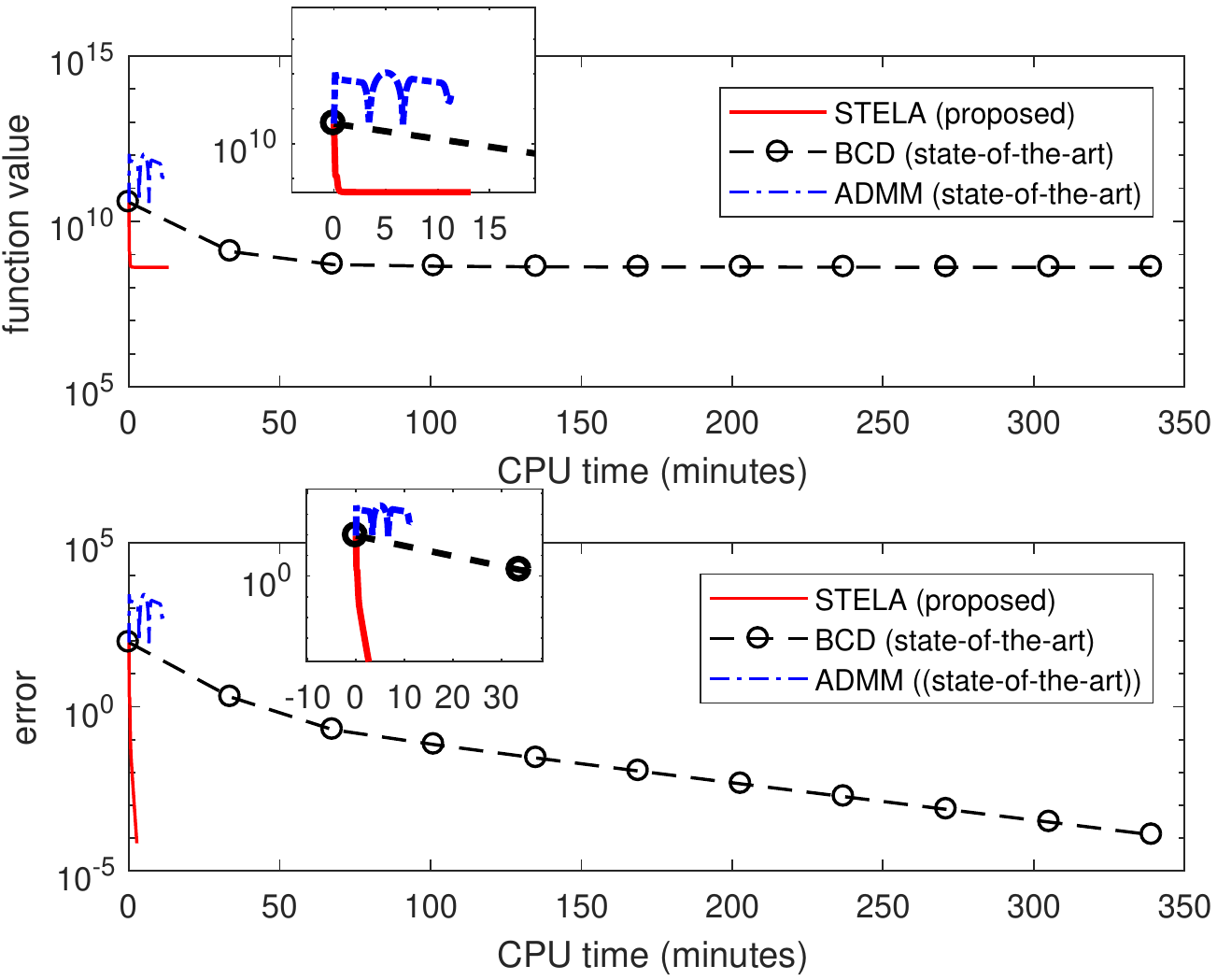}\caption{\label{fig:RankMinimization}Sparsity regularized rank minimization:
achieved function value $h(\mathbf{Z}^{t})$ versus the number of
iterations and CPU time (in minutes).}
\end{figure}

In this subsection, we perform numerical tests to compare the proposed
algorithm STELA with the BCD algorithm \cite{Mardani2013b} and the
ADMM algorithm \cite{Mardani2013}. We start with a brief description
of the ADMM algorithm: the problem (\ref{eq:rank-problem-formulation})
can be rewritten as
\begin{align}
\underset{\mathbf{P},\mathbf{Q},\mathbf{A},\mathbf{B}}{\textrm{minimize}}\quad & \frac{1}{2}\left\Vert \mathbf{P}\mathbf{Q}+\mathbf{D}\mathbf{A}-\mathbf{Y}\right\Vert _{F}^{2}+\frac{\lambda}{2}\left(\left\Vert \mathbf{P}\right\Vert _{F}^{2}+\left\Vert \mathbf{Q}\right\Vert _{F}^{2}\right)+\mu\left\Vert \mathbf{B}\right\Vert _{1}\nonumber \\
\textrm{subject to}\quad & \mathbf{A=B}.\label{eq:rank-problem-formulation-admm}
\end{align}
The augmented Lagrangian of (\ref{eq:rank-problem-formulation-admm})
is
\begin{align*}
L_{c}(\mathbf{P},\mathbf{Q},\mathbf{A},\mathbf{B},\boldsymbol{\Pi})= & \frac{1}{2}\left\Vert \mathbf{P}\mathbf{Q}+\mathbf{D}\mathbf{A}-\mathbf{Y}\right\Vert _{F}^{2}+\frac{\lambda}{2}\left(\left\Vert \mathbf{P}\right\Vert _{F}^{2}+\left\Vert \mathbf{Q}\right\Vert _{F}^{2}\right)\\
 & +\mu\left\Vert \mathbf{B}\right\Vert _{1}+\textrm{tr}(\boldsymbol{\Pi}^{T}(\mathbf{A-B}))+\frac{c}{2}\left\Vert \mathbf{A}-\mathbf{B}\right\Vert _{F}^{2},
\end{align*}
where $c$ is a positive constant. In ADMM, the variables are updated
in the $t$-th iteration as follows:
\begin{align*}
(\mathbf{Q}^{t+1},\mathbf{B}^{t+1}) & =\underset{\mathbf{Q},\mathbf{A}}{\arg\min}\;L_{c}(\mathbf{P}^{t},\mathbf{Q},\mathbf{A}^{t},\mathbf{B},\boldsymbol{\Pi}^{t}),\\
\mathbf{P}^{t+1} & =\underset{\mathbf{P}}{\arg\min}\;L_{c}(\mathbf{P},\mathbf{Q}^{t+1},\mathbf{A}^{t+1},\mathbf{B}^{t},\boldsymbol{\Pi}^{t}),\\
\mathbf{A}^{t+1} & =\underset{\mathbf{B}}{\arg\min}\;L_{c}(\mathbf{P}^{t+1},\mathbf{Q}^{t+1},\mathbf{A},\mathbf{B}^{t+1},\boldsymbol{\Pi}^{t}),\\
\boldsymbol{\Pi}^{t+1} & =\boldsymbol{\Pi}^{t}+c(\mathbf{A}^{t+1}-\mathbf{B}^{t+1}).
\end{align*}
Note that the solutions to the above optimization problems have an
analytical expression \cite{Mardani2013}. We set $c=10^{4}$.

The simulation parameters are set as follows. $N=1000$, $K=4000$,
$I=4000$, $\rho=10$. The elements of $\mathbf{D}$ are binary and
generated randomly and they are either 0 or 1. The elements of $\mathbf{V}$
follow the Gaussian distribution with mean 0 and variance $0.01$.
Each element of $\mathbf{S}$ can take three possible values, namely,
-1, 0,1, with the probability $P(S_{i,k}=-1)=P(S_{ik}=1)=0.05$ and
$P(S_{ik}=0)=0.9$. We set $\mathbf{Y}=\mathbf{PQ}+\mathbf{DS}+\mathbf{V}$,
where $\mathbf{P}$ and \textbf{$\mathbf{Q}$} are generated randomly
following the Gaussian distribution $\mathcal{N}(\mathbf{0},100/I)$
and $\mathcal{N}(\mathbf{0},100/K)$, respectively. The sparsity regularization
parameters are $\lambda=0.1\cdot\left\Vert \mathbf{Y}\right\Vert $
($\left\Vert \mathbf{Y}\right\Vert $ is the spectral norm of $\mathbf{Y}$)
and $\mu=0.1\cdot\left\Vert \mathbf{D}^{T}\mathbf{Y}\right\Vert _{\infty}$.
The simulation results are averaged over 20 realizations. For the
visual convenience, the curves of STELA and ADMM are magnified in
a small window inside the same figure.

In Fig. \ref{fig:RankMinimization} (a) and (b), we show respectively
the achieved objective function value and error versus the CPU time
(in minutes) by different algorithms, namely, STELA, BCD and ADMM.
In Fig. \ref{fig:RankMinimization} (b), the error is defined as $(f(\mathbf{Z}^{t})+g(\mathbf{S}^{t})-f(\mathbf{Z}^{\star})-g(\mathbf{S}^{\star}))/(f(\mathbf{Z}^{\star})+g(\mathbf{S}^{\star}))$,
where $\mathbf{Z}^{\star}$ is obtained by running the proposed algorithm
STELA for a sufficiently large number of iterations. As we see from
Fig. \ref{fig:RankMinimization} (a), the ADMM does not converge,
as the optimization problem (\ref{eq:rank-problem-formulation-admm})
(and (\ref{eq:rank-problem-formulation})) is nonconvex. We also observe
that the behavior of the ADMM is very sensitive to the value of $c$:
in some instances, the ADMM may converge if $c$ is large enough,
but it is a difficult task on its own to choose an appropriate value
of $c$ to achieve a good performance.

We run the BCD algorithm for 10 iterations, each represented by a
circle. In each iteration, all rows of $\mathbf{S}$ are updated once
in a sequential order, and it incurs a large delay. In particular,
we see from Fig. \ref{fig:RankMinimization} (a) that each iteration
of the BCD algorithm takes about 35 minutes, and a reasonably good
solution is obtained after two iterations (70 minutes). By contrast,
all variables are updated simultaneously in STELA and the CPU time
needed for each iteration is very small. We see from Fig. \ref{fig:RankMinimization}
(b) that STELA converges to a stationary point with a precision of
$10^{-5}$ in less than 1 minute, while it takes the BCD algorithm
about 330 minutes (5.5 hours) to find a solution that has the same
precision. This marks a notable improvement which is important in
real time anomaly detection in large networks.

\section{\label{sec:application-CappedL1}Sparse Subspace Clustering Through
Capped $\ell_{1}$-Norm Minimization}

In this section, we consider the sparse subspace clustering problem
through the capped $\ell_{1}$-norm minimization introduced in Sec.
\ref{sec:Problem-Formulation}-B:
\[
\underset{\mathbf{x}}{\textrm{minimize}}\quad\frac{1}{2}\left\Vert \mathbf{Ax-b}\right\Vert _{2}^{2}+\mu\sum_{k=1}^{K}\min(|x_{k}|,\theta),
\]
or more compactly,
\begin{equation}
\underset{\mathbf{x}}{\textrm{minimize}}\quad\frac{1}{2}\left\Vert \mathbf{Ax-b}\right\Vert _{2}^{2}+\mu\left\Vert \min(|\mathbf{x}|,\theta\mathbf{1})\right\Vert _{1}.\label{eq:capped-l1}
\end{equation}
It is shown in \cite{Gasso2009} that problem (\ref{eq:capped-l1})
is a special case of (\ref{eq:problem-formulation}) by setting\begin{subequations}
\begin{align}
f(\mathbf{x}) & \triangleq\frac{1}{2}\left\Vert \mathbf{Ax-b}\right\Vert _{2}^{2},\label{eq:capped-l1-f}\\
g^{+}(\mathbf{x}) & \triangleq\mu\left\Vert \mathbf{x}\right\Vert _{1},\label{eq:capped-l1-g+}\\
g^{-}(\mathbf{x}) & \triangleq\mu\left\Vert \mathbf{x}\right\Vert _{1}-\mu\left\Vert \min(|\mathbf{x}|,\theta\mathbf{1})\right\Vert _{1}.\label{eq:capped-l1-g-}
\end{align}
\end{subequations}

Since $f$ is convex, we adopt the best-response type approximate
function: the approximate function consists of $K$ component functions,
and in the $k$-th component function, only the $k$-th element, $x_{k}$,
of $\mathbf{x}$ is treated as a variable while other elements $\mathbf{x}_{-k}\triangleq(x_{j})_{j\neq k}$
are fixed,
\begin{equation}
\tilde{f}(\mathbf{x};\mathbf{x}^{t})=\frac{1}{2}\sum_{k=1}^{K}f(x_{k},\mathbf{x}_{-k}^{t})=\frac{1}{2}\sum_{k=1}^{K}\biggl\Vert\mathbf{a}_{k}x_{k}+\sum_{j\neq k}\mathbf{a}_{j}x_{j}^{t}-\mathbf{b}\biggr\Vert_{2}^{2}.\label{eq:capped-l1-approximate-function}
\end{equation}
To obtain the update direction, we solve the approximate problem
\begin{align}
\mathbb{B}\mathbf{x}^{t} & =\underset{\mathbf{x}}{\arg\min}\;\{\tilde{f}(\mathbf{x};\mathbf{x}^{t})-(\mathbf{x}-\mathbf{x}^{t})\boldsymbol{\xi}^{-}(\mathbf{x}^{t})+g^{+}(\mathbf{x})\}\nonumber \\
 & =\mathbf{d}(\mathbf{A}^{T}\mathbf{A})^{-1}\circ\mathcal{S}_{\mu\mathbf{1}}(\mathbf{r}(\mathbf{x}^{t},\boldsymbol{\xi}^{-}(\mathbf{x}^{t}))),\label{eq:capped-l1-best-response}
\end{align}
where
\begin{align*}
\mathbf{r}(\mathbf{x}^{t},\boldsymbol{\xi}^{-}(\mathbf{x}^{t})) & \triangleq\mathbf{d}(\nabla^{2}f(\mathbf{x}^{t}))\circ\mathbf{x}^{t}+\boldsymbol{\xi}^{-}(\mathbf{x}^{t})-\nabla f(\mathbf{x}^{t})\\
 & =\mathbf{d}(\mathbf{A}^{T}\mathbf{A})\circ\mathbf{x}^{t}+\boldsymbol{\xi}^{-}(\mathbf{x}^{t})-\mathbf{A}^{T}(\mathbf{A}\mathbf{x}^{t}-\mathbf{b}),
\end{align*}
$\mathbf{d}(\mathbf{X})$ is the diagonal vector of $\mathbf{X}$,
$\mathcal{S}_{\mathbf{a}}(\mathbf{b})\triangleq[\mathbf{b-a}]^{+}-[\mathbf{-b-a}]^{+}$
is the soft-thresholding operator, and the subgradient of $g^{-}(\mathbf{x})$
defined in (\ref{eq:rank-approximate-function-individual}) is $\boldsymbol{\xi}^{-}(\mathbf{x})=(\xi_{k}^{-}(x_{k}))_{k=1}^{K}$
with
\[
\xi_{k}^{-}(x_{k})=\begin{cases}
\mu, & \textrm{if }x_{k}\ge\theta,\\
-\mu, & \textrm{if }x_{k}\leq-\theta,\\
0, & \textrm{otherwise},
\end{cases}
\]
or more compactly,
\[
\boldsymbol{\xi}^{-}(\mathbf{x})=\frac{1}{2}\mu(\textrm{sign}(\mathbf{x}-\boldsymbol{\theta})-\textrm{sign}(-\mathbf{x}-\boldsymbol{\theta})).
\]

Given the update direction $\mathbb{B}\mathbf{x}^{t}-\mathbf{x}^{t}$,
we calculate the stepsize $\gamma^{t}$ according to the proposed
exact line search (\ref{eq:exact-line-search-proposed}), which can
be performed in a simple closed-form expression:
\begin{align}
\gamma^{t} & =\underset{0\leq\gamma\leq1}{\arg\min}\left\{ \negthickspace\negmedspace\begin{array}{l}
f(\mathbf{x}^{t}+\gamma(\mathbb{B}\mathbf{x}^{t}-\mathbf{x}^{t}))\smallskip\\
+\gamma(g^{+}(\mathbb{B}\mathbf{x}^{t})-g^{+}(\mathbf{x}^{t})-(\mathbb{B}\mathbf{x}^{t}-\mathbf{x}^{t})^{T}\boldsymbol{\xi}^{-}(\mathbf{x}^{t}))
\end{array}\negthickspace\negmedspace\right\} \nonumber \\
 & \negthickspace\negthickspace=\left[\frac{(\boldsymbol{\xi}^{-}(\mathbf{x}^{t})\negmedspace-\negmedspace\mathbf{A}^{T}(\mathbf{A}\mathbf{x}^{t}\negmedspace-\negmedspace\mathbf{b}))^{T}(\mathbb{B}\mathbf{x}^{t}\negmedspace-\negmedspace\mathbf{x}^{t})\negmedspace-\negmedspace\mu(\left\Vert \mathbb{B}\mathbf{x}^{t}\right\Vert _{1}\negmedspace-\negmedspace\left\Vert \mathbf{x}^{t}\right\Vert _{1})}{(\mathbf{A}(\mathbb{B}\mathbf{x}^{t}-\mathbf{x}^{t}))^{T}(\mathbf{A}(\mathbb{B}\mathbf{x}^{t}-\mathbf{x}^{t}))}\right]_{\negmedspace0}^{\negmedspace1}.\label{eq:capped-l1-stepsize}
\end{align}
The proposed update (\ref{eq:capped-l1-best-response})-(\ref{eq:capped-l1-stepsize})
are summarized in Algorithm \ref{alg:capped-L1} and we name it as
Soft-Thresholding with Exact Line search Algorithm (STELA). It has
several attractive features:

\begin{algorithm}[t]
\textbf{Data: }$t=0$, $\mathbf{x}^{0}$ (arbitrary but fixed, e.g.,
$\mathbf{x}^{0}=\mathbf{0}$), stop criterion $\delta$.

\textbf{S1: }Compute $\mathbb{B}\mathbf{x}^{t}$ according to (\ref{eq:best-response}).

\textbf{S2: }Determine the stepsize $\gamma^{t}$ by the exact line
search (\ref{eq:exact-line-search-proposed}).

\textbf{S3: }Update\textbf{ $\mathbf{x}^{t+1}$} according to (\ref{eq:variable-update}).

\textbf{S4: }If $|(\mathbb{B}\mathbf{x}^{t}-\mathbf{x}^{t})^{T}(\nabla f(\mathbf{x}^{t})-\boldsymbol{\xi}^{-}(\mathbf{x}^{t}))+g^{+}(\mathbb{B}\mathbf{x}^{t})-g^{+}(\mathbf{x}^{t})|\leq\delta$,
STOP; otherwise $t\leftarrow t+1$ and go to \textbf{S1}.

\caption{\label{alg:capped-L1}STELA: The proposed parallel best-response with
exact line search algorithm for the capped $\ell_{1}$-norm minimization
problem (\ref{eq:capped-l1})}
\end{algorithm}
\begin{itemize}
\item \emph{i) low complexity}, as the approximate function is chosen such
that its minimum can be obtained in closed-form expressions and the
proposed algorithm thus has a single layer. Besides this, the stepsize
can also be computed by closed-form expressions;
\item \emph{ii) fast convergence}, as all elements are updated in parallel,
the approximate function is of a best-response type, and the stepsize
is based on the exact line search;
\item \emph{iii) guaranteed convergence}, as $\tilde{f}(\mathbf{x};\mathbf{x}^{t})$
in (\ref{eq:capped-l1-approximate-function}) is strongly convex and
Assumptions (A4)-(A5) are satisfied.
\end{itemize}
Compared with state-of-the-art algorithms proposed for problem (\ref{eq:capped-l1}),
we remark that
\begin{itemize}
\item feature i) is an advantage over the traditional MM method \cite{Gasso2009};
\item feature ii) is an advantage over the algorithms \cite{Gong2013,Attouch2013}
with a proximal type approximation;
\item feature iii) is an advantage over the standard SCA framework for convex
regularization functions \cite{Yang_ConvexApprox,Scutari_BigData,Razaviyayn2014}.
\end{itemize}
\textbf{On the comparison with the proximal MM method \cite{Gong2013}.}
The proximal type algorithm proposed in \cite{Gong2013} is essentially
a MM method, because the variable is updated by
\begin{equation}
\mathbf{x}^{t+1}=\underset{\mathbf{x}\in\mathcal{X}}{\arg\min}\negthickspace\left\{ \negthickspace\begin{array}{l}
f(\mathbf{x}^{t})+\nabla f(\mathbf{x}^{t})^{T}(\mathbf{x}-\mathbf{x}^{t})+\frac{c^{t}}{2}\left\Vert \mathbf{x}-\mathbf{x}^{t}\right\Vert ^{2}\smallskip\\
+g^{+}(\mathbf{x})-g^{-}(\mathbf{x})
\end{array}\negthickspace\right\} ,\label{eq:cappedL1-MM}
\end{equation}
with $c^{t}>L_{\nabla f}$, while the objective function in (\ref{eq:cappedL1-MM})
is a global upper bound of $h(\mathbf{x})$ in view of the descent
lemma \cite[Prop. A.24]{bertsekas1999nonlinear}. When the value of
$L_{\nabla f}$ is not known, $c^{t}$ is estimated iteratively: for
some constants $0<\alpha<1$ and $0<\beta<1$, set $\mathbf{x}^{t+1}=\mathbf{x}^{\star}(\beta^{m_{t}})$
, where $\mathbf{x}^{\star}(\beta^{m})$ is defined as
\begin{equation}
\mathbf{x}^{\star}(\beta^{m})\negthickspace\triangleq\negthickspace\underset{\mathbf{x}}{\arg\min}\left\{ \negthickspace\negthickspace\begin{array}{l}
f(\mathbf{x}^{t})+\nabla f(\mathbf{x}^{t})^{T}(\mathbf{x}-\mathbf{x}^{t})+\frac{1}{2\beta^{m}}\left\Vert \mathbf{x}-\mathbf{x}^{t}\right\Vert ^{2}\smallskip\\
+g^{+}(\mathbf{x})-g^{-}(\mathbf{x})
\end{array}\negthickspace\negthickspace\right\} \label{eq:proximal-gradient-line-search}
\end{equation}
and $m_{t}$ is the smallest nonnegative integer such that $h(\mathbf{x}^{\star}(\beta^{m_{t}}))-h(\mathbf{x}^{t})\leq-\alpha/2\beta^{m_{t}}\left\Vert \mathbf{x}^{\star}(\beta^{m_{t}})-\mathbf{x}^{t}\right\Vert ^{2}$.
As a result, $\mathbf{x}^{\star}(\beta^{m})$ must be evaluated repeatedly
for $m_{t}$ times, namely, $m=0,1,\ldots,m_{t}$. This is however
not necessary in the proposed algorithm STELA, because computing the
descent direction and the stepsize according to (\ref{eq:capped-l1-best-response})
and (\ref{eq:capped-l1-stepsize}) does not depend on any unknown
parameters. Furthermore, (\ref{eq:proximal-gradient-line-search})
may not be easy to solve for a general $g^{-}(\mathbf{x})$ except
for some specific choices studied in \cite{Gong2013}.

\begin{figure}[t]
\includegraphics[scale=0.65]{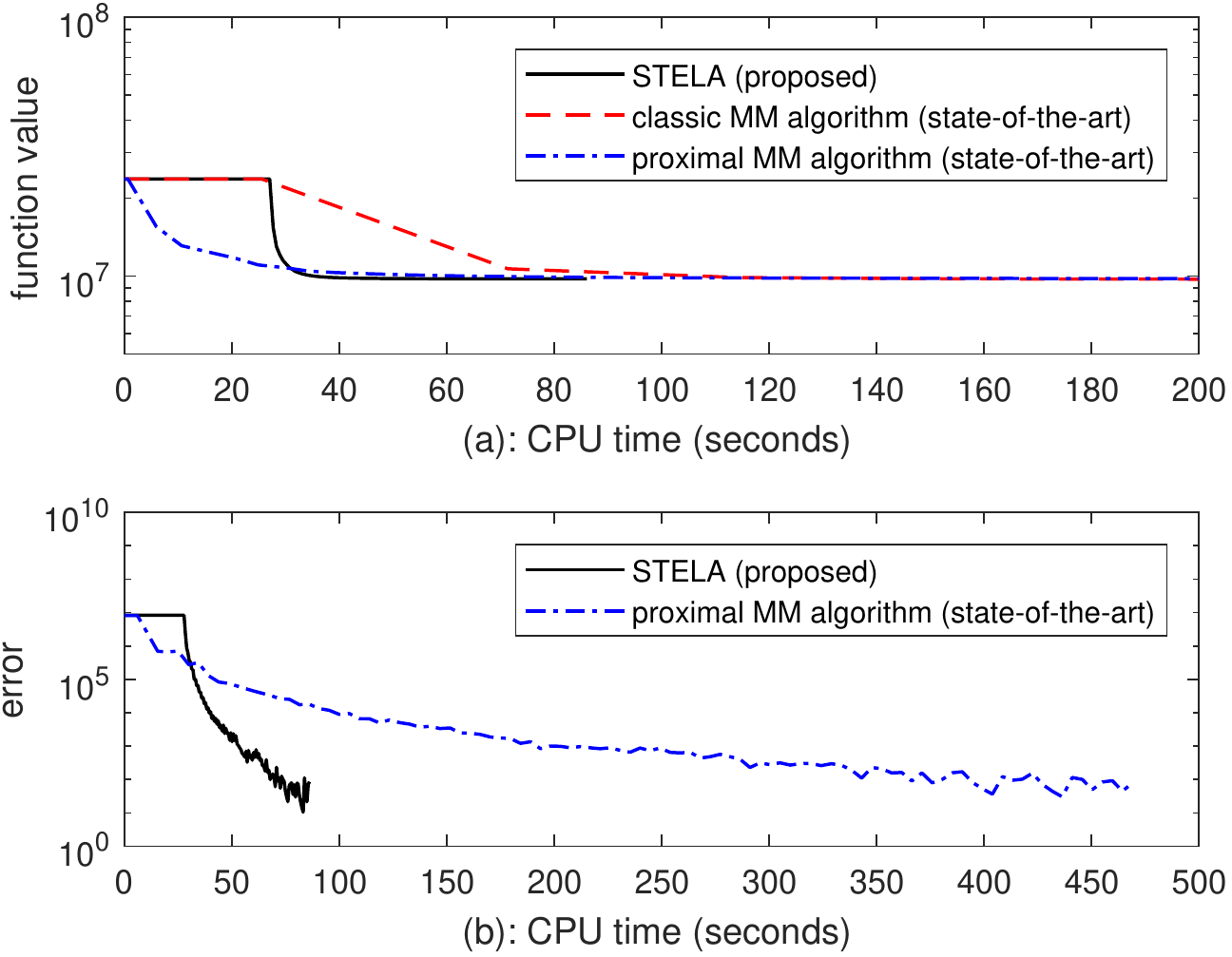}\caption{\label{fig:Capped--norm-minimization}Capped $\ell_{1}$-norm minimization:
achieved function value $h(\mathbf{x}^{t})$ and error $h(\mathbf{x}^{t-1})-h(\mathbf{x}^{t})$
versus CPU time (in seconds).}
\end{figure}

\subsection{Numerical Simulations}

In our numerical simulations the dimension of $\mathbf{A}$ is $10000\times50000$:
all of its elements are generated randomly by the normal distribution
$\mathcal{N}(0,1)$, and the rows of $\mathbf{A}$ are normalized
to have a unit $\ell_{2}$-norm. The density (the proportion of nonzero
elements) of the sparse vector $\mathbf{x}_{\textrm{true}}$ is 0.1.
The vector $\mathbf{b}$ is generated as $\mathbf{b}=\mathbf{A}\mathbf{x}_{\textrm{true}}+\mathbf{e}$
where $\mathbf{e}$ is drawn from an i.i.d. Gaussian distribution
with mean 0 and variance $10^{-4}$. The regularization parameter
$\mu$ is set to $\mu=0.1\left\Vert \mathbf{A}^{T}\mathbf{b}\right\Vert _{\infty}$,
which allows $\mathbf{x}_{\textrm{true}}$ to be recovered to a high
accuracy \cite{Wright2009}, and the parameter $\theta$ in the capped
$\ell_{1}$-norm is set to 1.

We compare the proposed algorithm STELA with the classic MM method
\cite{Gasso2009} and the proximal MM algorithm \cite{Gong2013}.
The comparison is made in terms of CPU time that is required until
the maximum number of iterations (100 for STELA and the proximal MM
algorithm and 10 for the classic MM method) is reached. The running
time consists of both the initialization stage required for preprocessing
(represented by a flat curve) and the formal stage in which the iterations
are carried out. For example, in STELA, $\mathbf{d}(\mathbf{A}^{T}\mathbf{A})$
is computed in the initialization stage since it is required in the
iterative variable update in the formal stage, cf. (\ref{eq:capped-l1-best-response}).
The upper bound function in the classic MM method, cf. (\ref{eq:CCCP}),
is minimized by STELA for $\ell_{1}$-norm (with a warm start that
sets the optimal point of the previous iteration as the initial point
of the current iteration \cite[Sec. II-D]{Gasso2009}), which was
presented in \cite[Sec. IV-III]{Yang_ConvexApprox}. All algorithms
have the same initial point, $\mathbf{x}^{0}=\mathbf{0}$. The simulation
results are averaged over 20 instances.

The achieved function value $h(\mathbf{x}^{t})$ and error $h(\mathbf{x}^{t-1})-h(\mathbf{x}^{t})$
versus the CPU time (in seconds) is plotted in Fig. \ref{fig:Capped--norm-minimization}
(a) and \ref{fig:Capped--norm-minimization} (b), respectively. We
see from Fig. \ref{fig:Capped--norm-minimization} (a) that all algorithms
converge to the same value. Furthermore, the initialization stage
of STELA is much longer than that of the proximal MM algorithm, because
computing $\mathbf{d}(\mathbf{A}^{T}\mathbf{A})$, the diagonal vector
of $\mathbf{A}^{T}\mathbf{A}$, is computationally expensive, especially
when the dimension of $\mathbf{A}$ is large. Nevertheless, in the
formal stage, the convergence speed of STELA is much faster than the
proximal MM algorithm, and this is mainly due to the use of the best-response
type approximate function (\ref{eq:capped-l1-approximate-function}),
and more specifically, the use of $\mathbf{d}(\mathbf{A}^{T}\mathbf{A})$,
cf. (\ref{eq:capped-l1-best-response}), which represents partial
second order information of the function $f$ in (\ref{eq:capped-l1})
(note that $\nabla^{2}f(\mathbf{x})=\mathbf{A}^{T}\mathbf{A}$). We
see from Fig. \ref{fig:Capped--norm-minimization} (b) that the long
initialization stage is compensated by the fast convergence speed
in the formal stage. We mention for the paper's completeness that
$\mathbf{d}(\mathbf{A}^{T}\mathbf{A})$ can be calculated analytically
in some applications, e.g., when $\mathbf{A}$ is a Vandermonde or
constant modulus matrix.

We see from Fig. \ref{fig:Capped--norm-minimization} (a) that the
major complexity of the classic MM method lies in the first few iterations,
as the complexity of late iterations are notably reduced by a good
initialization thanks to the warm start. The most notable difference
between the MM method and the STELA is that the upper bound function
is only approximately minimized in the STELA, and this leads to a
significant reduction in the computational complexity. Using the approximate
function is also beneficial when the upper bound function $\overline{h}(\mathbf{x};\mathbf{x}^{t})$
is not easy to minimize, e.g., $f(\mathbf{x})$ is nonconvex.

\section{\label{sec:Concluding-Remarks}Concluding Remarks}

In this paper, we have proposed a successive convex approximation
framework for sparse signal estimation where the nonsmooth nonconvex
regularization function is nonconvex and can be written as the difference
of two convex functions. The proposed procedure is to apply the standard
successive convex approximation for convex regularization functions
to an upper bound of the original objective function that can be obtained
following the standard MM method. This procedure also facilitates
the design of low-complexity line search schemes which are carried
out over a differentiable function. The proposed framework is flexible
and it leads to algorithms that exploit the problem structure and
have a low complexity. Customizing the general framework for the example
applications in network anomaly detection and sparse subspace clustering,
the proposed algorithm STELA is a best-response type algorithm with
exact line search and it has several attractive features, illustrated
both theoretically and numerically: i) fast convergence due to the
best-response type approximation and the line search for stepsize
calculation; ii) low complexity as both the optimal point of the approximate
function and the exact line search have closed-form expressions; and
iii) guaranteed convergence to a stationary point.

\appendices{}

\section{\label{sec:Proof-of-Propositions}Proof of Propositions \ref{prop:descent-property}
and \ref{prop:stepsize}}
\begin{IEEEproof}
[Proof of Proposition \ref{prop:descent-property}]Since the approximate
problem (\ref{eq:approximate-problem}) is convex, $\mathbb{B}\mathbf{x}^{t}$
is a globally optimal point of (\ref{eq:approximate-problem}) and
\[
\tilde{h}(\mathbb{B}\mathbf{x}^{t};\mathbf{x}^{t})=\min_{\mathbf{x}\in\mathcal{X}}\tilde{h}(\mathbf{x};\mathbf{x}^{t})\leq\tilde{h}(\mathbf{x}^{t};\mathbf{x}^{t}).
\]
We discuss the two possibilities separately, namely,
\begin{equation}
\textrm{i): }\tilde{h}(\mathbb{B}\mathbf{x}^{t};\mathbf{x}^{t})=\tilde{h}(\mathbf{x}^{t};\mathbf{x}^{t}),\label{eq:possibility-1}
\end{equation}
or
\begin{equation}
\textrm{ii): }\tilde{h}(\mathbb{B}\mathbf{x}^{t};\mathbf{x}^{t})<\tilde{h}(\mathbf{x}^{t};\mathbf{x}^{t}).\label{eq:possibility-2}
\end{equation}

i) $\tilde{h}(\mathbb{B}\mathbf{x}^{t};\mathbf{x}^{t})=\tilde{h}(\mathbf{x}^{t};\mathbf{x}^{t})$.
We show that $\tilde{h}(\mathbb{B}\mathbf{x}^{t};\mathbf{x}^{t})=\tilde{h}(\mathbf{x}^{t};\mathbf{x}^{t})$
is equivalent to $\mathbf{x}^{t}$ being a stationary point of (\ref{eq:problem-formulation}).

If $\tilde{h}(\mathbb{B}\mathbf{x}^{t};\mathbf{x}^{t})=\tilde{h}(\mathbf{x}^{t};\mathbf{x}^{t})$,
then $\mathbf{x}^{t}\in\mathcal{S}(\mathbf{x}^{t})$:
\[
\mathbf{x}^{t}\in\underset{\mathbf{x}\in\mathcal{X}}{\arg\min}\;\tilde{h}(\mathbf{x};\mathbf{x}^{t}),
\]
and it must satisfy the first-order optimality condition: for some
$\boldsymbol{\xi}^{+}(\mathbf{x}^{t})$,
\[
(\mathbf{x}-\mathbf{x}^{t})(\nabla\tilde{f}(\mathbf{x}^{t};\mathbf{x}^{t})+\boldsymbol{\xi}^{+}(\mathbf{x}^{t})-\boldsymbol{\xi}^{-}(\mathbf{x}^{t}))\geq0,\forall\mathbf{x}.
\]
This is exactly the first-order optimality condition of problem (\ref{eq:problem-formulation})
after replacing $\nabla\tilde{f}(\mathbf{x}^{t};\mathbf{x}^{t})$
by $\nabla f(\mathbf{x}^{t})$ in view of Assumption (A3) on the gradient
consistency. Therefore, $\mathbf{x}^{t}$ is a stationary point of
(\ref{eq:problem-formulation}).

Reversely, if $\mathbf{x}^{t}$ is a stationary point of (\ref{eq:problem-formulation}),
then it satisfies the first-order optimality condition: for some $\boldsymbol{\xi}^{+}(\mathbf{x}^{t})$
and $\boldsymbol{\xi}^{-}(\mathbf{x}^{t})$,
\[
(\mathbf{x}-\mathbf{x}^{t})(\nabla f(\mathbf{x}^{t})+\boldsymbol{\xi}^{+}(\mathbf{x}^{t})-\boldsymbol{\xi}^{-}(\mathbf{x}^{t}))\geq0,\forall\mathbf{x}.
\]
By assumption (A3) on the gradient consistency, the above condition
is equivalent to
\[
(\mathbf{x}-\mathbf{x}^{t})(\nabla\tilde{f}(\mathbf{x}^{t};\mathbf{x}^{t})+\boldsymbol{\xi}^{+}(\mathbf{x}^{t})-\boldsymbol{\xi}^{-}(\mathbf{x}^{t}))\geq0,\forall\mathbf{x}.
\]
Since problem (\ref{eq:approximate-problem}) is convex, the above
condition implies that $\mathbf{x}^{t}$ is a globally optimal point
of (\ref{eq:best-response}) and $\tilde{h}(\mathbf{x}^{t};\mathbf{x}^{t})=\min_{\mathbf{x}\in\mathcal{X}}\tilde{h}(\mathbf{x};\mathbf{x}^{t})$.

ii) $\tilde{h}(\mathbb{B}\mathbf{x}^{t};\mathbf{x}^{t})<\tilde{h}(\mathbf{x}^{t};\mathbf{x}^{t})$.
We remark that problem (\ref{eq:approximate-problem}) is convex and
equivalent to the following problem
\begin{align}
\underset{\mathbf{x},y}{\textrm{minimize}}\quad & \tilde{f}(\mathbf{x};\mathbf{x}^{t})-(\mathbf{x}-\mathbf{x}^{t})^{T}\boldsymbol{\xi}^{-}(\mathbf{x}^{t})+y\nonumber \\
\textrm{subject to}\quad & \mathbf{x}\in\mathcal{X},g^{+}(\mathbf{x})\leq y.\label{eq:approximate-problem-differentiable}
\end{align}
The equivalence between (\ref{eq:approximate-function}) and (\ref{eq:approximate-problem-differentiable})
is in the sense that $\mathbb{B}\mathbf{x}^{t}$ defined in (\ref{eq:best-response})
is the optimal $\mathbf{x}$ of (\ref{eq:approximate-problem-differentiable}),
and the optimal $y$ of (\ref{eq:approximate-problem-differentiable}),
denoted as $y^{\star}(\mathbf{x}^{t})$, is given by $y^{\star}(\mathbf{x}^{t})=g^{+}(\mathbb{B}\mathbf{x}^{t})$.
If $\tilde{h}(\mathbb{B}\mathbf{x}^{t};\mathbf{x}^{t})<\tilde{h}(\mathbf{x}^{t};\mathbf{x}^{t})$,
then
\begin{align*}
\tilde{h}(\mathbb{B}\mathbf{x}^{t};\mathbf{x}^{t}) & =\tilde{f}(\mathbb{B}\mathbf{x}^{t};\mathbf{x}^{t})-(\mathbb{B}\mathbf{x}^{t}-\mathbf{x}^{t})^{T}\boldsymbol{\xi}^{-}(\mathbf{x}^{t})+g^{+}(\mathbb{B}\mathbf{x}^{t})\\
 & \overset{(a)}{=}\tilde{f}(\mathbb{B}\mathbf{x}^{t};\mathbf{x}^{t})-(\mathbb{B}\mathbf{x}^{t}-\mathbf{x}^{t})^{T}\boldsymbol{\xi}^{-}(\mathbf{x}^{t})+y^{\star}(\mathbf{x}^{t})\\
 & <\tilde{h}(\mathbf{x}^{t};\mathbf{x}^{t})\\
 & =\tilde{f}(\mathbf{x}^{t};\mathbf{x}^{t})-(\mathbf{x}^{t}-\mathbf{x}^{t})^{T}\boldsymbol{\xi}^{-}(\mathbf{x}^{t})+g^{+}(\mathbf{x}^{t})\\
 & \overset{(b)}{\leq}\tilde{f}(\mathbf{x}^{t};\mathbf{x}^{t})-(\mathbf{x}^{t}-\mathbf{x}^{t})^{T}\boldsymbol{\xi}^{-}(\mathbf{x}^{t})+y^{t},
\end{align*}
where the equality (\emph{a}) follows from the fact that $y^{\star}(\mathbf{x}^{t})=g^{+}(\mathbb{B}\mathbf{x}^{t})$,
and the inequality (\emph{b}) follows from the fact that $y^{t}\geq g^{+}(\mathbf{x}^{t})$
in view of the constraint in (\ref{eq:approximate-problem-differentiable}).
Since $y^{t}$ does not appear in (\ref{eq:approximate-problem-differentiable}),
we set without loss of generality $y^{t}=g^{+}(\mathbf{x}^{t})$.

The objective function of (\ref{eq:approximate-problem-differentiable})
is convex and differentiable, and thus also pseudoconvex \cite[Figure 1]{Yang_ConvexApprox}.
From the definition of pseudoconvex functions that
\begin{align*}
 & \tilde{f}(\mathbb{B}\mathbf{x}^{t};\mathbf{x}^{t})-(\mathbb{B}\mathbf{x}^{t}-\mathbf{x}^{t})^{T}\boldsymbol{\xi}^{-}(\mathbf{x}^{t})+y^{\star}(\mathbf{x}^{t})\\
<\; & \tilde{f}(\mathbf{x}^{t};\mathbf{x}^{t})-(\mathbf{x}^{t}-\mathbf{x}^{t})^{T}\boldsymbol{\xi}^{-}(\mathbf{x}^{t})+y^{t},
\end{align*}
implies
\[
(\mathbb{B}\mathbf{x}^{t}-\mathbf{x}^{t})^{T}(\nabla\tilde{f}(\mathbf{x}^{t};\mathbf{x}^{t})-\boldsymbol{\xi}^{-}(\mathbf{x}^{t}))+y^{\star}(\mathbb{B}\mathbf{x}^{t})-y^{t}<0,
\]
which is equivalent to the following inequality after replacing $\nabla\tilde{f}(\mathbf{x}^{t};\mathbf{x}^{t})$
by $\nabla f(\mathbf{x}^{t})$ in view of Assumption (A3) on the gradient
consistency:
\begin{equation}
(\mathbb{B}\mathbf{x}^{t}-\mathbf{x}^{t})^{T}(\nabla f(\mathbf{x}^{t})-\boldsymbol{\xi}^{-}(\mathbf{x}^{t}))+y^{\star}(\mathbb{B}\mathbf{x}^{t})-y^{t}<0,\label{eq:descent-direction-differentiable}
\end{equation}
where $y^{\star}(\mathbb{B}\mathbf{x}^{t})=g^{+}(\mathbb{B}\mathbf{x}^{t})$
and $y^{t}=g^{+}(\mathbf{x}^{t})$. Therefore, we readily obtain the
inequality in (\ref{eq:descent-direction}) and the proof of Proposition
\ref{prop:descent-property} is thus completed.
\end{IEEEproof}
\begin{IEEEproof}
[Proof of Proposition \ref{prop:stepsize}]We define
\[
l(\mathbf{x},y;\mathbf{x}^{t})\triangleq f(\mathbf{x})-(\mathbf{x}-\mathbf{x}^{t})^{T}\boldsymbol{\xi}^{-}(\mathbf{x}^{t})+y.
\]
We can see that $\nabla_{\mathbf{x}}l(\mathbf{x},y;\mathbf{x}^{t})=\nabla f(\mathbf{x})-\boldsymbol{\xi}^{-}(\mathbf{x}^{t})$
and $\nabla_{y}l(\mathbf{x},y;\mathbf{x}^{t})=1$. Then the inequality
(\ref{eq:descent-direction-differentiable}) can be rewritten as
\begin{align*}
0 & >(\mathbb{B}\mathbf{x}^{t}-\mathbf{x}^{t})^{T}\nabla_{\mathbf{x}}l(\mathbf{x}^{t},y^{t};\mathbf{x}^{t})+(y^{\star}(\mathbb{B}\mathbf{x}^{t})-y^{t})\nabla_{y}l(\mathbf{x}^{t},y^{t};\mathbf{x}^{t})\\
 & =(\mathbb{B}\mathbf{x}^{t}-\mathbf{x}^{t},y^{\star}(\mathbf{x}^{t})-y^{t})^{T}\nabla l(\mathbf{x}^{t},y^{t};\mathbf{x}^{t})
\end{align*}
From the above inequality we can claim that $(\mathbb{B}\mathbf{x}^{t},y^{\star}(\mathbf{x}^{t}))-(\mathbf{x}^{t},y^{t})$
is a descent direction of the function $l(\mathbf{x},y;\mathbf{x}^{t})=f(\mathbf{x})-(\mathbf{x}-\mathbf{x}^{t})^{T}\boldsymbol{\xi}^{-}(\mathbf{x}^{t})+y$
at the point $(\mathbf{x}^{t},y^{t})$.

The proposed exact line search (\ref{eq:exact-line-search-proposed})
is equivalent to applying the standard exact line search to the differentiable
function $l(\mathbf{x},y;\mathbf{x}^{t})$ along the direction $(\mathbb{B}\mathbf{x}^{t},y^{\star}(\mathbf{x}^{t}))-(\mathbf{x}^{t},y^{t})$:
\[
\gamma^{t}=\underset{0\leq\gamma\leq1}{\arg\min}\left\{ \begin{array}{l}
f(\mathbf{x}^{t}+\gamma(\mathbb{B}\mathbf{x}^{t}-\mathbf{x}^{t}))\smallskip\\
-(\mathbf{x}^{t}+\gamma(\mathbb{B}\mathbf{x}^{t}-\mathbf{x}^{t})-\mathbf{x}^{t})^{T}\boldsymbol{\xi}^{-}(\mathbf{x}^{t})\smallskip\\
+y^{t}+\gamma(y^{\star}(\mathbf{x}^{t})-y^{t}).
\end{array}\right\} .
\]
Therefore the existence of a $\gamma^{t}\in(0,1]$ is guaranteed according
to \cite[8.2.1]{Ortega&Rheinboldt}.

Similarly, the proposed successive line search is equivalent to applying
the standard successive line search to the differentiable function
$l(\mathbf{x},y;\mathbf{x}^{t})$ along the direction $(\mathbb{B}\mathbf{x}^{t},y^{\star}(\mathbf{x}^{t}))-(\mathbf{x}^{t},y^{t})$:
\begin{align*}
 & l(\mathbf{x}^{t}+\beta^{m}(\mathbb{B}\mathbf{x}^{t}-\mathbf{x}^{t}),y^{t}+\beta^{m}(y^{\star}(\mathbf{x}^{t})-y^{t});\mathbf{x}^{t})\\
\leq\; & l(\mathbf{x}^{t},y^{t};\mathbf{x}^{t})+\alpha\beta^{m}(\mathbb{B}\mathbf{x}^{t}-\mathbf{x}^{t},y^{\star}(\mathbf{x}^{t})-y^{t})^{T}\nabla l(\mathbf{x}^{t},y^{t};\mathbf{x}^{t}).
\end{align*}
The proof of Proposition \ref{prop:stepsize} is thus completed.
\end{IEEEproof}

\section{\label{sec:Proof-of-Theorem}Proof of Theorem \ref{thm:convergence}}
\begin{IEEEproof}
Similar to \cite[Theorem 1]{Yang_ConvexApprox}, the key of the proof
is to show that $\mathbb{B}\mathbf{x}$ is a closed mapping \cite{Berge1963},
i.e., if $\lim_{t\rightarrow\infty}\mathbf{x}^{t}=\mathbf{x}$ and
$\lim_{t\rightarrow\infty}\mathbb{B}\mathbf{x}^{t}=\mathbf{y}$, then
$\mathbb{B}\mathbf{x}\in\mathcal{S}(\mathbf{x})$. The key difference
is that the objective function $h$ in (\ref{eq:problem-formulation})
is nondifferentiable.

Since $\mathbb{B}\mathbf{x}^{t}$ is the optimal point of (\ref{eq:approximate-problem}),
it satisfies the first-order optimality condition:
\begin{equation}
(\mathbf{x}-\mathbb{B}\mathbf{x}^{t})^{T}(\nabla\tilde{f}(\mathbb{B}\mathbf{x}^{t};\mathbf{x}^{t})-\boldsymbol{\xi}^{-}(\mathbf{x}^{t})+\boldsymbol{\xi}^{+}(\mathbf{x}^{t}))\geq0,\;\forall\,\mathbf{x}\in\mathcal{X}.\label{eq:minimum-principle}
\end{equation}

If (\ref{eq:possibility-1}) is true, then $\mathbf{x}^{t}\in\mathcal{S}(\mathbf{x}^{t})$
and it is a stationary point of (\ref{eq:problem-formulation}) according
to Proposition \ref{prop:descent-property} (i). Besides, it follows
from (\ref{eq:problem-formulation}) (with $\mathbf{x}=\mathbb{B}\mathbf{x}^{t}$
and $\mathbf{y}=\mathbf{x}^{t}$) that $(\mathbb{B}\mathbf{x}^{t}-\mathbf{x}^{t})^{T}(\nabla f(\mathbf{x}^{t})-\boldsymbol{\xi}^{-}(\mathbf{x}^{t})+\boldsymbol{\xi}^{+}(\mathbf{x}^{t}))\geq0$.
Note that equality is actually achieved, i.e.,
\[
(\mathbb{B}\mathbf{x}^{t}-\mathbf{x}^{t})^{T}(\nabla f(\mathbf{x}^{t})-\boldsymbol{\xi}^{-}(\mathbf{x}^{t})+\boldsymbol{\xi}^{+}(\mathbf{x}^{t}))=0
\]
because otherwise $\mathbb{B}\mathbf{x}^{t}-\mathbf{x}^{t}$ would
be an ascent direction of $\tilde{h}(\mathbf{x};\mathbf{x}^{t})$
at $\mathbf{x}=\mathbf{x}^{t}$ and the definition of $\mathbb{B}\mathbf{x}^{t}$
would be contradicted. Then from the definition of the proposed successive
line search in (\ref{eq:successive-line-search-proposed}), we can
readily infer that
\begin{equation}
h(\mathbf{x}^{t+1})\leq h(\mathbf{x}^{t}).\label{eq:decreasing-1}
\end{equation}
It is easy to see (\ref{eq:decreasing-1}) holds for the exact line
search as well.

If (\ref{eq:possibility-2}) is true, $\mathbf{x}^{t}$ is not a stationary
point and $\mathbb{B}\mathbf{x}^{t}-\mathbf{x}^{t}$ is a strict descent
direction of $h(\mathbf{x})$ at $\mathbf{x}=\mathbf{x}^{t}$ according
to Proposition \ref{prop:descent-property} (ii): $h(\mathbf{x})$
is strictly decreased compared with $h(\mathbf{x}^{t})$ if $\mathbf{x}$
is updated at $\mathbf{x}^{t}$ along the direction $\mathbb{B}\mathbf{x}^{t}-\mathbf{x}^{t}$.
From Proposition \ref{prop:stepsize}, the proposed successive line
search schemes yield a stepsize $\gamma^{t}$ such that $0<\gamma^{t}\leq1$
and
\begin{equation}
h(\mathbf{x}^{t+1})=h(\mathbf{x}^{t}+\gamma^{t}(\mathbb{B}\mathbf{x}^{t}-\mathbf{x}^{t}))<h(\mathbf{x}^{t}).\label{eq:decreasing-2}
\end{equation}
This strict decreasing property also holds for the exact line search
because it is the stepsize that yields the largest decrease, which
is always larger than or equal to that of the successive line search.

We know from (\ref{eq:decreasing-1}) and (\ref{eq:decreasing-2})
that $\left\{ h(\mathbf{x}^{t})\right\} $ is a monotonically decreasing
sequence and it thus converges. Besides, for any two (possibly different)
convergent subsequences $\left\{ \mathbf{x}^{t}\right\} _{t\in\mathcal{T}_{1}}$
and $\left\{ \mathbf{x}^{t}\right\} _{t\in\mathcal{T}_{2}}$, the
following holds:
\[
\lim_{t\rightarrow\infty}h(\mathbf{x}^{t})=\lim_{\mathcal{T}_{1}\ni t\rightarrow\infty}h(\mathbf{x}^{t})=\lim_{\mathcal{T}_{2}\ni t\rightarrow\infty}h(\mathbf{x}^{t}).
\]
Since $h(\mathbf{x})$ is a continuous function, we infer from the
preceding equation that
\begin{equation}
h\left(\lim_{\mathcal{T}_{1}\ni t\rightarrow\infty}\mathbf{x}^{t}\right)=h\left(\lim_{\mathcal{T}_{2}\ni t\rightarrow\infty}\mathbf{x}^{t}\right).\label{eq:value-convergence}
\end{equation}

Now consider any convergent subsequence $\{\mathbf{x}^{t}\}_{t\in\mathcal{T}}$
with limit point $\mathbf{y}$, i.e., $\lim_{\mathcal{T}\ni t\rightarrow\infty}\mathbf{x}^{t}=\mathbf{y}$.
To show that $\mathbf{y}$ is a stationary point, we first assume
the contrary: $\mathbf{y}$ is not a stationary point. Since $\tilde{h}(\mathbf{x};\mathbf{x}^{t})$
is continuous in both $\mathbf{x}$ and $\mathbf{x}^{t}$ by Assumption
(A2) and $\left\{ \mathbf{B}\mathbf{x}^{t}\right\} _{t\in\mathcal{T}}$
is bounded by Assumption (A5), there exists a sequence $\left\{ \mathbb{B}\mathbf{x}^{t}\right\} _{t\in\mathcal{T}_{s}}$
with $\mathcal{T}_{s}\subseteq\mathcal{T}$ such that it converges
and it follows from the Maximum Theorem in \cite[Ch. VI.3]{Berge1963}
that $\lim_{\mathcal{T}_{s}\ni t\rightarrow\infty}\mathbb{B}\mathbf{x}^{t}\in\mathcal{S}(\mathbf{y})$.
Since both $f(\mathbf{x})$ and $\nabla f(\mathbf{x})$ are continuous,
applying the Maximum Theorem again implies there is a $\mathcal{T}_{s'}$
such that $\mathcal{T}_{s'}\subseteq\mathcal{T}_{s}(\subseteq\mathcal{T})$
and $\left\{ \mathbf{x}^{t+1}\right\} _{t\in\mathcal{T}_{s'}}$ converges
to $\mathbf{y}'$ defined as $\mathbf{y}'\triangleq\mathbf{y}+\rho(\mathbb{B}\mathbf{y}-\mathbf{y})$,
where $\rho$ is the stepsize when either the exact or successive
line search is applied to $f(\mathbf{y})$ along the direction $\mathbb{B}\mathbf{y}-\mathbf{y}$.
Since $\mathbf{y}$ is not a stationary point, it follows from (\ref{eq:decreasing-2})
that $h(\mathbf{y}')<h(\mathbf{y})$, but this would contradict (\ref{eq:value-convergence}).
Therefore $\mathbf{y}$ is a stationary point, and the proof is completed.
\end{IEEEproof}


\begin{thebibliography}{10}
\providecommand{\url}[1]{#1}
\csname url@samestyle\endcsname
\providecommand{\newblock}{\relax}
\providecommand{\bibinfo}[2]{#2}
\providecommand{\BIBentrySTDinterwordspacing}{\spaceskip=0pt\relax}
\providecommand{\BIBentryALTinterwordstretchfactor}{4}
\providecommand{\BIBentryALTinterwordspacing}{\spaceskip=\fontdimen2\font plus
\BIBentryALTinterwordstretchfactor\fontdimen3\font minus
  \fontdimen4\font\relax}
\providecommand{\BIBforeignlanguage}[2]{{%
\expandafter\ifx\csname l@#1\endcsname\relax
\typeout{** WARNING: IEEEtran.bst: No hyphenation pattern has been}%
\typeout{** loaded for the language `#1'. Using the pattern for}%
\typeout{** the default language instead.}%
\else
\language=\csname l@#1\endcsname
\fi
#2}}
\providecommand{\BIBdecl}{\relax}
\BIBdecl

\bibitem{Theodoridis_2015}
S.~Theodoridis, \emph{{Machine Learning: A Bayesian and Optimization
  Perspective}}, 1st~ed.\hskip 1em plus 0.5em minus 0.4em\relax Academic Press,
  2015.

\bibitem{Beck2009a}
A.~Beck and M.~Teboulle, ``{A Fast Iterative Shrinkage-Thresholding
  Algorithm},'' \emph{Society for Industrial and Applied Mathematics Journal on
  Imaging Sciences}, vol.~2, no.~1, pp. 183--202, 2009.

\bibitem{Tseng2001}
P.~Tseng, ``{Convergence of a Block Coordinate Descent Method for
  Nondifferentiable Minimization},'' \emph{Journal of Optimization Theory and
  Applications}, vol. 109, no.~3, pp. 475--494, Jun. 2001.

\bibitem{Boyd2010}
\BIBentryALTinterwordspacing
S.~Boyd, N.~Parikh, E.~Chu, B.~Peleato, and J.~Eckstein, ``{Distributed
  Optimization and Statistical Learning via the Alternating Direction Method of
  Multipliers},'' \emph{Foundations and Trends in Machine Learning}, vol.~3,
  no.~1, 2010.
\BIBentrySTDinterwordspacing

\bibitem{Parikh2014}
\BIBentryALTinterwordspacing
N.~Parikh and S.~Boyd, ``{Proximal Algorithms},'' \emph{Foundations and Trends
  in Optimization}, vol.~1, no.~3, pp. 127--239, 2014.
\BIBentrySTDinterwordspacing

\bibitem{Yang_ConvexApprox}
Y.~Yang and M.~Pesavento, ``{A Unified Successive Pseudoconvex Approximation
  Framework},'' \emph{IEEE Transactions on Signal Processing}, vol.~65, no.~13,
  pp. 3313--3328, Jul. 2017.

\bibitem{Yang2015_arxiv_ICML}
\BIBentryALTinterwordspacing
Z.~Yang, Z.~Wang, H.~Liu, Y.~C. Eldar, and T.~Zhang, ``{Sparse Nonlinear
  Regression: Parameter Estimation and Asymptotic Inference},'' in
  \emph{International Conference on Machine Learning (ICML)}, 2016.
\BIBentrySTDinterwordspacing

\bibitem{Bertsekas}
D.~P. Bertsekas and J.~N. Tsitsiklis, \emph{{Parallel and distributed
  computation: Numerical methods}}.\hskip 1em plus 0.5em minus 0.4em\relax
  Prentice Hall, 1989.

\bibitem{Razaviyayn2013}
\BIBentryALTinterwordspacing
M.~Razaviyayn, M.~Hong, and Z.-Q. Luo, ``{A Unified Convergence Analysis of
  Block Successive Minimization Methods for Nonsmooth Optimization},''
  \emph{SIAM Journal on Optimization}, vol.~23, no.~2, pp. 1126--1153, Jan.
  2013.
\BIBentrySTDinterwordspacing

\bibitem{Beck2013}
\BIBentryALTinterwordspacing
A.~Beck and L.~Tetruashvili, ``{On the Convergence of Block Coordinate Descent
  Type Methods},'' \emph{SIAM Journal on Optimization}, vol.~23, no.~4, pp.
  2037--2060, Jan. 2013.
\BIBentrySTDinterwordspacing

\bibitem{Wright2015}
\BIBentryALTinterwordspacing
S.~J. Wright, ``{Coordinate descent algorithms},'' \emph{Mathematical
  Programming}, vol. 151, no.~1, pp. 3--34, 2015.
\BIBentrySTDinterwordspacing

\bibitem{Mardani2013b}
M.~Mardani, G.~Mateos, and G.~B. Giannakis, ``{Dynamic anomalography: Tracking
  network anomalies via sparsity and low rank},'' \emph{IEEE Journal on
  Selected Topics in Signal Processing}, vol.~7, no.~1, pp. 50--66, Feb. 2013.

\bibitem{Slavakis2014a}
\BIBentryALTinterwordspacing
K.~Slavakis, G.~B. Giannakis, and G.~Mateos, ``{Modeling and Optimization for
  Big Data Analytics: (Statistical) learning tools for our era of data
  deluge},'' \emph{IEEE Signal Processing Magazine}, vol.~31, no.~5, pp.
  18--31, Sep. 2014.
\BIBentrySTDinterwordspacing

\bibitem{Elad2006}
M.~Elad, ``{Why simple shrinkage is still relevant for redundant
  representations?}'' \emph{IEEE Transactions on Information Theory}, vol.~52,
  no.~12, pp. 5559--5569, Dec. 2006.

\bibitem{Razaviyayn2014}
\BIBentryALTinterwordspacing
M.~Razaviyayn, M.~Hong, Z.-Q. Luo, and J.-S. Pang, ``{Parallel Successive
  Convex Approximation for Nonsmooth Nonconvex Optimization},'' in
  \emph{Proceedings of the 27th International Conference on Neural Information
  Processing Systems}, 2014, pp. 1440--1448.
\BIBentrySTDinterwordspacing

\bibitem{Scutari_BigData}
\BIBentryALTinterwordspacing
F.~Facchinei, G.~Scutari, and S.~Sagratella, ``{Parallel Selective Algorithms
  for Nonconvex Big Data Optimization},'' \emph{IEEE Transactions on Signal
  Processing}, vol.~63, no.~7, pp. 1874--1889, Nov. 2015.
\BIBentrySTDinterwordspacing

\bibitem{Steffens2016}
C.~Steffens, Y.~Yang, and M.~Pesavento, ``{Multidimensional sparse recovery for
  MIMO channel parameter estimation},'' \emph{European Signal Processing
  Conference}, pp. 66--70, 2016.

\bibitem{Mardani2013}
M.~Mardani, G.~Mateos, and G.~B. Giannakis, ``{Decentralized
  sparsity-regularized rank minimization: Algorithms and applications},''
  \emph{IEEE Transactions on Signal Processing}, vol.~61, no.~21, pp.
  5374--5388, Nov. 2013.

\bibitem{Hong2016}
\BIBentryALTinterwordspacing
M.~Hong, Z.-Q. Luo, and M.~Razaviyayn, ``{Convergence Analysis of Alternating
  Direction Method of Multipliers for a Family of Nonconvex Problems},''
  \emph{SIAM Journal on Optimization}, vol.~26, no.~1, pp. 337--364, Jan. 2016.
\BIBentrySTDinterwordspacing

\bibitem{Jiang2016}
\BIBentryALTinterwordspacing
B.~Jiang, T.~Lin, S.~Ma, and S.~Zhang, ``{Structured Nonconvex and Nonsmooth
  Optimization: Algorithms and Iteration Complexity Analysis},'' 2016.
  [Online]. Available: \url{http://arxiv.org/abs/1605.02408}
\BIBentrySTDinterwordspacing

\bibitem{Tibshirani2011a}
\BIBentryALTinterwordspacing
R.~Tibshirani, ``{Regression shrinkage and selection via the lasso: a
  retrospective},'' \emph{Journal of the Royal Statistical Society: Series B
  (Statistical Methodology)}, vol.~73, no.~3, pp. 273--282, Jun. 2011.
\BIBentrySTDinterwordspacing

\bibitem{Fan2001}
\BIBentryALTinterwordspacing
J.~Fan and R.~Li, ``{Variable Selection via Nonconcave Penalized Likelihood and
  its Oracle Properties},'' \emph{Journal of the American Statistical
  Association}, vol.~96, no. 456, pp. 1348--1360, Dec. 2001.
\BIBentrySTDinterwordspacing

\bibitem{Candes2008a}
\BIBentryALTinterwordspacing
E.~J. Cand{\`{e}}s, M.~B. Wakin, and S.~P. Boyd, ``{Enhancing Sparsity by
  Reweighted L1 Minimization},'' \emph{Journal of Fourier Analysis and
  Applications}, vol.~14, no. 5-6, pp. 877--905, Dec. 2008.
\BIBentrySTDinterwordspacing

\bibitem{Zhang2010}
T.~Zhang, ``{Analysis of Multi-stage Convex Relaxation for Sparse
  Regularization},'' \emph{Journal of Machine Learning Research}, vol.~11, pp.
  1081--1107, 2010.

\bibitem{Weston2003}
J.~Weston, A.~Elisseeff, B.~Scholkopf, and M.~Tipping, ``{The use of zero-norm
  with linear models and kernel methods},'' \emph{Journal of Machine Learning
  Research}, vol.~3, pp. 1439--1461, 2003.

\bibitem{Gasso2009}
G.~Gasso, A.~Rakotomamonjy, and S.~Canu, ``{Recovering sparse signals with a
  certain family of nonconvex penalties and DC programming},'' \emph{IEEE
  Transactions on Signal Processing}, vol.~57, no.~12, pp. 4686--4698, Dec.
  2009.

\bibitem{Sun2017}
Y.~Sun, P.~Babu, and D.~P. Palomar, ``{Majorization-Minimization Algorithms in
  Signal Processing, Communications, and Machine Learning},'' \emph{IEEE
  Transactions on Signal Processing}, vol.~65, no.~3, pp. 794--816, Feb. 2017.

\bibitem{Gong2013}
\BIBentryALTinterwordspacing
P.~Gong, C.~Zhang, Z.~Lu, J.~Huang, and J.~Ye, ``{A General Iterative Shrinkage
  and Thresholding Algorithm for Non-convex Regularized Optimization
  Problems},'' in \emph{Proceedings of the 30th International Conference on
  Machine Learning}, 2013, pp. 37--45.
\BIBentrySTDinterwordspacing

\bibitem{Attouch2013}
H.~Attouch, J.~Bolte, and B.~F. Svaiter, ``{Convergence of descent methods for
  semi-algebraic and tame problems: Proximal algorithms, forward-backward
  splitting, and regularized Gauss-Seidel methods},'' \emph{Mathematical
  Programming}, vol. 137, no. 1-2, pp. 91--129, 2013.

\bibitem{Yang_Rank_ICASSP2018}
\BIBentryALTinterwordspacing
Y.~Yang and M.~Pesavento, ``A parallel best-response algorithm with exact line
  search for nonconvex sparsity-regularized rank minimization,'' Apr. 2018, to
  appear in Proc. ICASSP. [Online]. Available:
  \url{http://orbilu.uni.lu/handle/10993/33772}
\BIBentrySTDinterwordspacing

\bibitem{Yang_MM_SCA_SAM2018}
\BIBentryALTinterwordspacing
Y.~Yang, M.~Pesavento, S.~Chatzinotas, and B.~Ottersten, ``Successive convex
  approximation algorithms for sparse signal estimation with nonconvex
  regularizations,'' 2018, technical report. [Online]. Available:
  \url{http://orbilu.uni.lu/handle/10993/35100}
\BIBentrySTDinterwordspacing

\bibitem{Burer2003}
\BIBentryALTinterwordspacing
S.~Burer and R.~D. Monteiro, ``{A nonlinear programming algorithm for solving
  semidefinite programs via low-rank factorization},'' \emph{Mathematical
  Programming}, vol.~95, no.~2, pp. 329--357, Feb. 2003.
\BIBentrySTDinterwordspacing

\bibitem{Recht2010}
\BIBentryALTinterwordspacing
B.~Recht, M.~Fazel, and P.~A. Parrilo, ``{Guaranteed Minimum-Rank Solutions of
  Linear Matrix Equations via Nuclear Norm Minimization},'' \emph{SIAM Review},
  vol.~52, no.~3, pp. 471--501, Jan. 2010.
\BIBentrySTDinterwordspacing

\bibitem{Elhamifar2013}
E.~Elhamifar and R.~Vidal, ``{Sparse Subspace Clustering: Algorithm, Theory,
  and Applications},'' \emph{IEEE Transactions on Pattern Analysis and Machine
  Intelligence}, vol.~35, no.~11, pp. 2765--2781, Nov. 2013.

\bibitem{bertsekas1999nonlinear}
D.~P. Bertsekas, \emph{{Nonlinear programming}}.\hskip 1em plus 0.5em minus
  0.4em\relax Athena Scientific, 1999.

\bibitem{Robinson1974}
\BIBentryALTinterwordspacing
S.~M. Robinson and R.~H. Day, ``{A sufficient condition for continuity of
  optimal sets in mathematical programming},'' \emph{Journal of Mathematical
  Analysis and Applications}, vol.~45, no.~2, pp. 506--511, Feb. 1974.
\BIBentrySTDinterwordspacing

\bibitem{Yang2015_arxiv}
\BIBentryALTinterwordspacing
Z.~Yang, Z.~Wang, H.~Liu, Y.~C. Eldar, and T.~Zhang, ``{Sparse Nonlinear
  Regression: Parameter Estimation and Asymptotic Inference},'' 2016, in Proc.
  International Conference on Machine Learning (ICML). [Online]. Available:
  \url{http://proceedings.mlr.press/v48/yangc16.pdf}
\BIBentrySTDinterwordspacing

\bibitem{Wright2009}
\BIBentryALTinterwordspacing
S.~Wright, R.~Nowak, and M.~Figueiredo, ``{Sparse Reconstruction by Separable
  Approximation},'' \emph{IEEE Transactions on Signal Processing}, vol.~57,
  no.~7, pp. 2479--2493, Jul. 2009.
\BIBentrySTDinterwordspacing

\bibitem{Ortega&Rheinboldt}
J.~M. Ortega and W.~C. Rheinboldt, \emph{{Iterative solution of nonlinear
  equations in several variables}}.\hskip 1em plus 0.5em minus 0.4em\relax
  Academic, New York, 1970.

\bibitem{Berge1963}
C.~Berge, \emph{{Topological Spaces: Including a Treatment of Multi-Valued
  Functions, Vector Spaces and Convexity}}.\hskip 1em plus 0.5em minus
  0.4em\relax Dover Publications, 1997.

\end{thebibliography}
\end{document}